\pdfoutput=1
\documentclass[11pt]{article}

\usepackage[final]{acl}
\usepackage{times}
\usepackage{latexsym}
\usepackage[T1]{fontenc}
\usepackage[utf8]{inputenc}
\usepackage{microtype}
\usepackage{inconsolata}
\usepackage{graphicx}

\usepackage{url}            
\usepackage{booktabs}       
\usepackage{amsfonts}       
\usepackage{nicefrac}       
\usepackage{xcolor}         
\usepackage{amsmath, amssymb, overpic, textpos}
\usepackage{multirow}
\usepackage{colortbl}
\usepackage{tabulary}
\usepackage{algorithmicx}
\usepackage{algpseudocode}
\usepackage{listings}
\usepackage{multicol}
\usepackage{xspace}
\usepackage{wrapfig}
\usepackage{graphbox}
\usepackage{arydshln}
\usepackage{algorithm}
\usepackage{adjustbox}
\usepackage{kotex}
\usepackage{newtxmath}
\usepackage{caption}
\usepackage{subcaption}
\usepackage{setspace}
\usepackage{makecell}

\definecolor{citecolor}{HTML}{0071BC}
\definecolor{linkcolor}{HTML}{ED1C24}

\usepackage{amsmath,amsfonts,bm}

\def\eqref#1{equation~\ref{#1}}

\def\1{\bm{1}}

\def\vs{{\bm{s}}}

\DeclareMathAlphabet{\mathsfit}{\encodingdefault}{\sfdefault}{m}{sl}
\SetMathAlphabet{\mathsfit}{bold}{\encodingdefault}{\sfdefault}{bx}{n}

\usepackage[capitalize]{cleveref}
\crefname{section}{Sec.}{Secs.}
\Crefname{section}{Section}{Sections}
\Crefname{table}{Table}{Tables}
\crefname{table}{Tab.}{Tabs.}
\usepackage{pifont}

\newcommand*\rot{\rotatebox{90}}

\makeatletter
\DeclareRobustCommand\onedot{\futurelet\@let@token\@onedot}
\def\@onedot{\ifx\@let@token.\else.\null\fi\xspace}

\def\eg{\emph{e.g}\onedot} 
\def\ie{\emph{i.e}\onedot} 
 
 \def\vs{\emph{vs}\onedot}

\makeatother
\newlength\savewidth

\newcolumntype{x}[1]{>{\centering\arraybackslash}p{#1pt}}
\newcolumntype{y}[1]{>{\raggedright\arraybackslash}p{#1pt}}
\newcolumntype{z}[1]{>{\raggedleft\arraybackslash}p{#1pt}}

\newlength\secmargin
\newlength\subsecmargin
\newlength\subsubsecmargin
\newlength\paramargin
\newlength\abovetabcapmargin
\newlength\belowtabcapmargin
\newlength\abovefigcapmargin
\newlength\belowfigcapmargin
\definecolor{Gray}{gray}{0.9}

\definecolor{Green}{rgb}{0.2, 0.7, 0.1}
\definecolor{prompt_blue}{HTML}{1f78b4}
\definecolor{prompt_red}{HTML}{d45c43}
\definecolor{plus}{HTML}{0071bc}
\definecolor{minus}{RGB}{153,10,10}
\definecolor{SecondBest}{HTML}{E9F6EC} 
\definecolor{Best}{HTML}{C8E6C9} 

\newcommand{\up}{\bf \fontsize{10}{42} \color{plus}{$\uparrow$}}
\newcommand{\down}{\bf \fontsize{10}{42}\selectfont \color{minus}{$\downarrow$}}

\newcommand{\vcentered}[1]{\ensuremath{\vcenter{\hbox{#1}}}}
\newcommand{\Ours}{\textsc{AvisC}\xspace}
\newcommand{\OursFull}{Attentional Vision Calibration\xspace}
\newcommand{\OursFullAbb}{\OursFull(\Ours)\xspace}
\newcommand{\Emoji}{\vcentered{\includegraphics[height=1.6em]{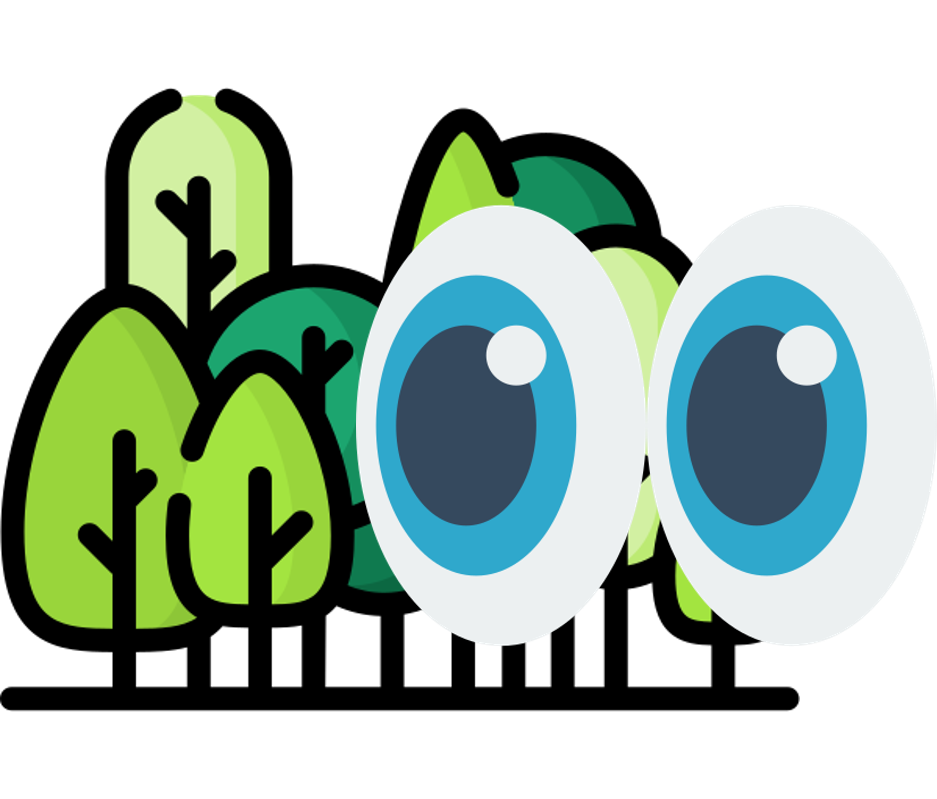}}\xspace}

\usepackage{hyperref}

\makeatletter
\let\oldthanks\thanks
\renewcommand{\thanks}[1]{%
  \begingroup
    \hypersetup{hidelinks}%
    \oldthanks{#1}%
  \endgroup
}
\makeatother

\title{\Emoji~Don't Miss the Forest for the Trees: \OursFull for Large Vision Language Models}

\author{
    Sangmin Woo\thanks{Equal contribution} \qquad
    Donguk Kim\footnotemark[1] \qquad
    Jaehyuk Jang\footnotemark[1] \qquad
    Yubin Choi \qquad
    Changick Kim\\
    KAIST \\
    {\tt\small \{smwoo95, kdu3613, jhyuk, choibinbin, changick\}@kaist.ac.kr}\\
    {\tt\textbf{Project: \url{https://sangminwoo.github.io/AvisC/}}}
}

\begin{document}
\maketitle
\begin{abstract}
Large Vision Language Models (LVLMs) demonstrate strong capabilities in visual understanding and description, yet often suffer from hallucinations, attributing incorrect or misleading features to images.
We observe that LVLMs disproportionately focus on a small subset of image tokens---termed \textit{blind tokens}---which are typically irrelevant to the query (\eg, background or non-object regions).
We hypothesize that such attention misalignment plays a key role in generating hallucinated responses.
To mitigate this issue, we propose \textbf{\OursFullAbb}, a test-time approach that dynamically recalibrates the influence of blind tokens without modifying the underlying attention mechanism.
\Ours first identifies blind tokens by analyzing layer-wise attention distributions over image tokens, then employs a contrastive decoding strategy to balance the influence of original and blind-token-biased logits.
Experiments on standard benchmarks, including POPE, MME, and AMBER, demonstrate that \Ours effectively reduces hallucinations in LVLMs.
\end{abstract}

\addtocontents{toc}{\protect\setcounter{tocdepth}{0}}
\vspace{\secmargin}\section{Introduction}\vspace{\secmargin}
\label{sec:intro}




Large Vision Language Models (LVLMs)~\citep{dai2024instructblip,zhu2023minigpt,liu2023visual,liu2023improved,bai2023qwen,tong2024cambrian} have recently shown remarkable capabilities in generating coherent and contextually relevant descriptions of visual inputs.
Yet, these models are prone to "hallucinations", producing responses that do not accurately reflect the underlying image.
Such hallucinations pose a critical challenge for applications, demanding reliability, precision, and trustworthy visual interpretation.



\begin{figure*}[t!]
    \centering
    \includegraphics[width=\textwidth]{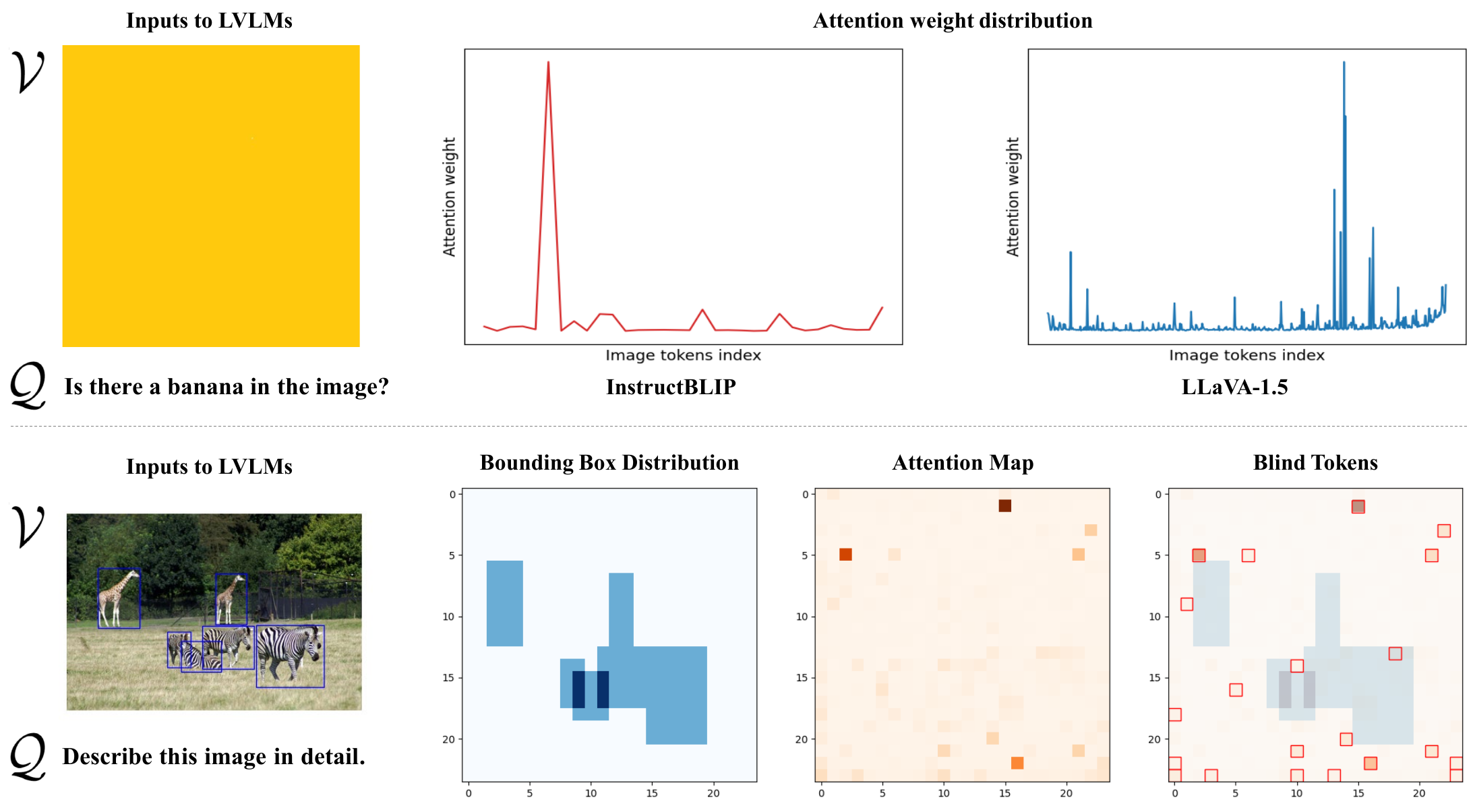}
    \vspace{\abovefigcapmargin}
    \vspace{-1mm}
    \caption{
    \textbf{Blind tokens in LVLMs.}
    \textbf{(Top)} Even when the image ($\mathcal{V}$) lacks meaningful content for the textual query ($\mathcal{Q})$, modern LVLMs~\citep{dai2024instructblip,liu2023visual} still assign disproportionate attention to a few image tokens (\ie, \textit{blind tokens}).
    Despite having identical, featureless yellow patches, these tokens dominate the attention distribution.
    \textbf{(Bottom)} In a real image, overlaying bounding boxes and LLaVA 1.5's attention map highlights a clear mismatch between blind tokens (red boxes) and genuinely informative regions.
    \textit{Note}: attention weights are averaged across all layers for the first generated token.
    See~\cref{sec:appendix_attention_bias,sec:appendix_statistics_blind_token} for more examples.
    }%
    \label{fig:observation_bias}
\end{figure*}

\begin{figure*}[t!]
    \centering
    \includegraphics[width=\textwidth]{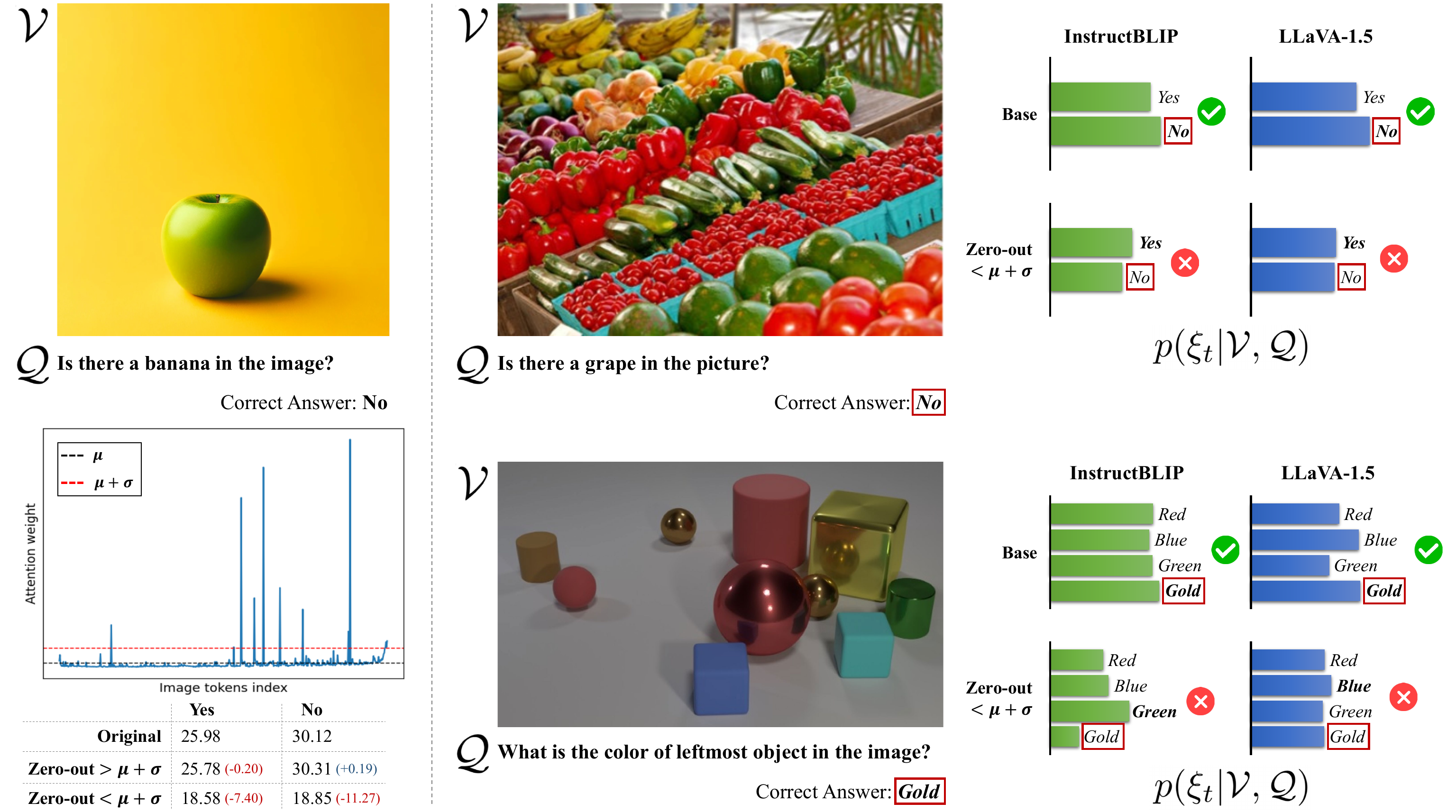}
    \vspace{\abovefigcapmargin}
    \vspace{-1mm}
    \caption{
        \textbf{Blind tokens contribute little to actual predictions.}
        \textbf{(a)}
        We perform zero-out experiments to measure the impact of blind \vs non-blind tokens. 
        Zeroing out blind tokens (Zero-out > $\mu + \sigma$), where attention weights are above mean + standard deviation, leaves the model’s predicted probabilities nearly unchanged, suggesting that these tokens carry minimal object-discriminative information.
        In contrast, zeroing out \textit{non-blind tokens} yields near 50:50 probabilities, underscoring their critical role in correct prediction.
        \textbf{(b)}
        When non-blind tokens are zeroed out, the models fails to correctly predict previously well-classified instances.
    }%
    \label{fig:motivation}
\end{figure*}

In this work, we hypothesize that a key contributing factor to hallucinations in LVLMs is an attention misalignment during inference. Specifically, we observe that LVLMs often allocate excessive attention to a small subset of image tokens—referred to as \textit{blind tokens}—which appear to be uninformative or irrelevant to the query (e.g., background or repetitive regions; see~\cref{fig:observation_bias}).
Our preliminary experiments support this observation (\cref{fig:motivation}): despite receiving high attention weights, these tokens do not typically carry the query-relevant information necessary for precise interpretation.


We posit that this skewed focus may cause LVLMs to rely disproportionately on such blind tokens during text generation, potentially leading to hallucinated or inaccurate responses. Rather than grounding their outputs in the most semantically relevant visual details, the models may be influenced by peripheral or misleading cues, undermining the factual integrity of their descriptions.


To investigate this hypothesis, we propose \Ours, a novel, training‐free decoding method that dynamically recalibrates the influence of blind tokens at inference time. \Ours operates in two key stages: first, it analyzes the layer-wise attention distributions to identify image tokens that receive excessive attention and flags them as blind tokens; then, it applies a contrastive decoding strategy~\citep{leng2023mitigating,favero2024multi}. By comparing the prediction logits generated using all image tokens with those obtained when blind tokens are selectively zeroed out, \Ours adjusts the final token probabilities to mitigate the undue influence of irrelevant tokens.

Notably, \Ours does not directly manipulate model's attention mechanism or require any retraining.
Instead, it operates solely at the decoding stage by modifying the final token probabilities, reducing the influence of blind tokens while amplifying the impact of non-blind tokens.


We validate our approach on multiple benchmarks---including POPE~\citep{rohrbach2018object}, MME~\citep{fu2024mme}, and AMBER~\citep{wang2023llm}---demonstrating that AvisC not only significantly reduces hallucinations but also enhances both the factual accuracy and descriptive richness of LVLM outputs. Importantly, our method is model-agnostic and requires no additional training data or external modules, making it a practical plug-and-play solution for improving existing LVLM systems.

In summary, our contribution is threefold:
\textbf{(1)} We uncover and analyze the phenomenon of \textit{blind tokens}---image tokens with disproportionately high attention that contribute to hallucinations in LVLMs.
\textbf{(2)} We introduce \Ours, a training‐free approach that dynamically recalibrates the influence of blind tokens without modifying the underlying model;
and \textbf{(3)} We demonstrate, through comprehensive experiments on standard benchmarks, that AvisC effectively reduces hallucinations and enhances the performance of diverse LVLMs.

\vspace{\secmargin}\section{Observations}\vspace{\secmargin}
\label{sec:observations}

Modern LVLMs~\citep{dai2024instructblip,liu2023visual} build upon the transformer architecture~\citep{vaswani2017attention}, where attention weights are intended to highlight the most relevant tokens for generating the next output token. Intuitively, tokens receiving higher attention should correspond to key elements in the image---an idea that has proven effective in both vision- and text-based transformers~\citep{caron2021emerging,ilharco2021openclip,vaswani2017attention}. However, we find that this principle does not always hold in current LVLMs.

\paragraph{Blind tokens in uniform images.}
A striking illustration of this issue arises when an image has no meaningful content for the query---such as a uniformly colored background. As shown in \cref{fig:observation_bias} (top), even in a plain yellow image with no discernible objects, LVLMs often concentrate most of their attention on a few patches. We refer to these excessively attended yet semantically uninformative patches as \emph{blind tokens}. This phenomenon echoes findings in vision transformers~\citep{darcet2023vision}\footnote{We discuss our findings in relation to~\citep{darcet2023vision} in~\cref{sec:appendix_discussion}.}, where certain background regions disproportionately attract high attention, possibly serving as global information “reservoirs” at the expense of local, detail-rich areas.

\paragraph{Mismatch between blind tokens and actual objects.}
Beyond artificially simple images, we also analyze real images from COCO2014~\citep{lin2014microsoft}. We ask LVLMs to describe the given image and measure how much attention goes to patches corresponding to object bounding boxes. As shown in \cref{fig:observation_bias} (bottom), many tokens receiving disproportionately high attention (blind tokens) have little overlap with genuine object regions, while actual objects receive comparatively less attention.
Specifically, only 3.7\% of blind tokens overlap with these regions, and merely 23.2\% of total attention weight is goes to them.\footnote{Detailed statistics can be found in~\cref{fig:blind_token_statistics}.} This mismatch indicates that, despite carrying little or no query-relevant information, blind tokens consume a large share of the model’s attention.
Consequently, truly informative tokens that capture critical visual details are underemphasized, potentially compromising the model’s descriptive quality and reliability.

\paragraph{Zero‐out experiments.}
To better understand the functional role of blind tokens, we conduct a zero‐out analysis on LLaVA‐1.5~\citep{liu2023visual}, shown in \cref{fig:motivation}. Specifically, we either zero out the blind tokens or the non‐blind tokens, then observe changes in the model’s predicted logits. When we remove blind tokens, the logits remain almost identical to those of the original model---indicating that these tokens contribute little to the final prediction. By contrast, removing non‐blind tokens causes the logits to collapse to near-uniform probabilities, revealing that the essential, object‐discriminative information resides in those less‐attended tokens.\footnote{For dataset-level zero‐out experiment details, see~\cref{tab:zero_out_experiments}.}

\paragraph{Attention bias and hallucinations.}
These findings suggest that LVLMs systematically overemphasize certain patches that do not meaningfully aid the prediction process. Consequently, truly informative tokens---often corresponding to the actual objects or key details---receive insufficient attention. We hypothesize that this imbalance predisposes the model to hallucinate, as the generation process leans on blind tokens that fail to encode crucial visual details. In the next sections, we propose a simple yet effective decoding method to mitigate this problem by recalibrating the model’s attention usage at decoding stage, thereby reducing its reliance on blind tokens and improving visual grounding.

\paragraph{Hypothesis.}
Our hypothesis is that blind tokens arises as a structural byproduct of the deep, layered architecture of these models, similar to the “high-norm outlier tokens” observed in vision transformers~\cite{darcet2023vision} (see~\cref{sec:appendix_discussion}).
As information is propagated through layers, global representations from earlier layers are progressively compressed.
However, instead of being allocated to semantically meaningful tokens, this global information often becomes concentrated in structurally convenient but semantically irrelevant tokens—frequently those in repetitive or low-information regions.
These tokens consequently accumulate disproportionate representational weight and attract excessive attention during decoding.
Despite their lack of semantic relevance, they misguide the model’s focus and contribute to the generation of hallucinated content.
While establishing a definitive causal link is inherently challenging, our  qualitative and quantitative evidence suggest that blind tokens are a recurring and impactful phenomenon in LVLMs.
Visualizations and token distribution analyses (see~\cref{fig:more_attention_bias,fig:blind_token_pope,fig:dist_heat_map,fig:blind_token_statistics,fig:histogram,fig:blind_token_prob}) demonstrate that blind tokens frequently emerge in spatially uninformative regions and exhibit anomalously high attention scores.

\vspace{\secmargin}\section{Approach: \Ours}
\begin{figure}[t!]
    \centering
    \vspace{-1mm}
    \includegraphics[width=\linewidth]{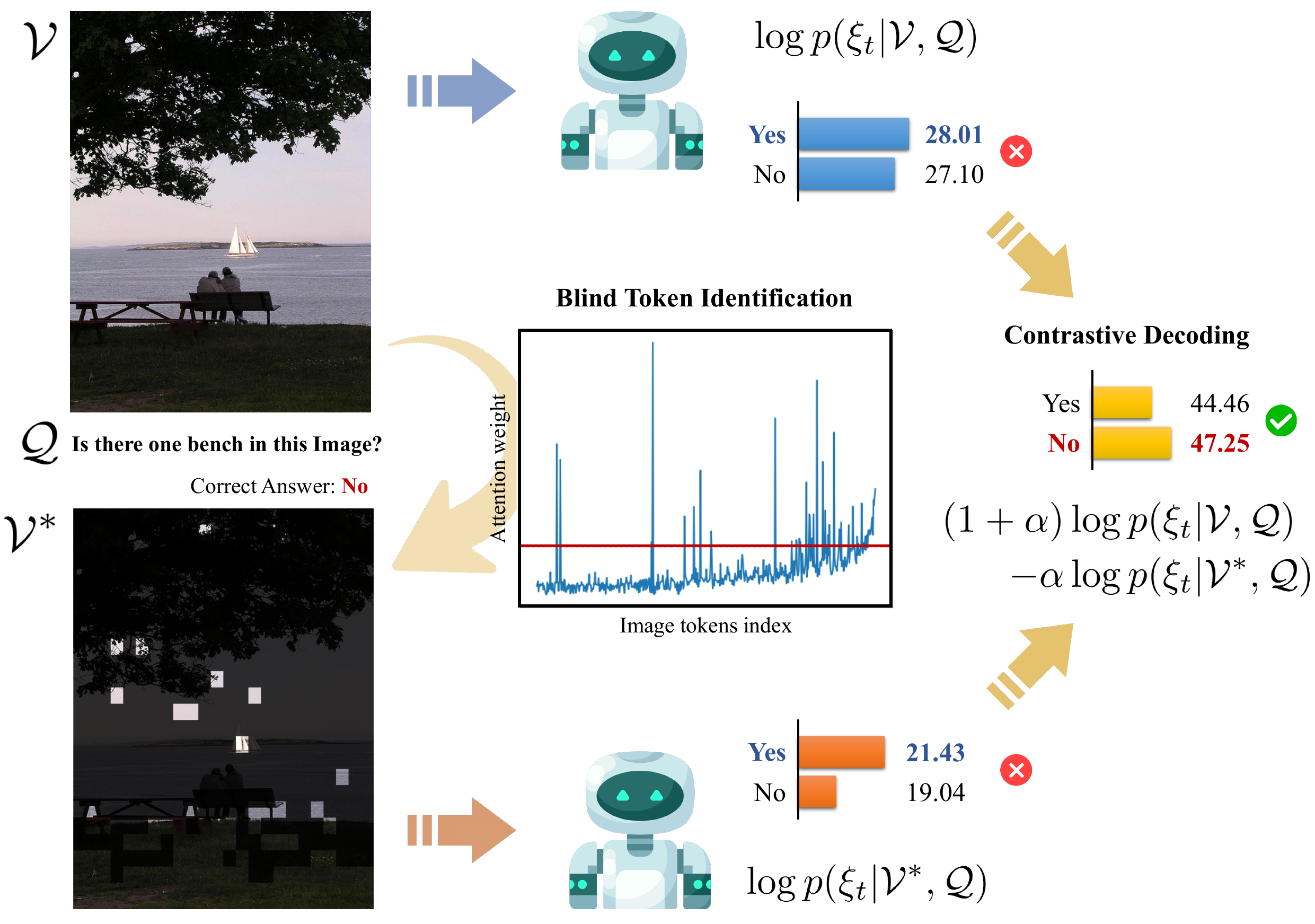}
    \vspace{\abovefigcapmargin}
    \caption{
    \textbf{An overview of \Ours.}
    }%
    \label{fig:overview}
\end{figure}

\vspace{\secmargin}
\label{sec:approach}
We propose \Ours, a test-time approach to enhance visual object understanding in LVLMs during decoding.
\Ours dynamically recalibrates the model’s attention at every token generation stepby reducing the over-emphasis on \emph{blind tokens}---image tokens that receive disproportionate attention despite lacking task-relevant information.
An overview of \Ours is shown in \cref{fig:overview}.
Our approach modifies the decoding process in three key steps: (1) \textbf{Layer selection}: identify layers that exhibit a high proportion of image-related attention; (2) \textbf{Blind token identification}: detect tokens that capture an unusually high share of attention; and (3) \textbf{Contrastive decoding}: adjust output logits to mitigate the influence of these blind tokens.


\vspace{\subsecmargin}\subsection{LVLM Framework}

\paragraph{Uni-modal encoding.}
LVLMs process visual inputs and textual queries into compact representations. Pre-trained encoders like CLIP~\citep{radford2021learning} are commonly used for processing visual data. The text data is tokenized, turning it into a sequence of manageable pieces for further processing.

\paragraph{Cross-modal alignment.}
Since LLMs natively process only text, a learnable cross-modal alignment module---such as the Q-Former~\citep{li2023blip} or a linear projection layer~\citep{liu2023visual}---transform visual features into tokens compatible with the LLM’s input space. This yields a set of visual tokens, $\mathcal{V} = \{\nu_0, \nu_1, \dots, \nu_{N-1}\}$, which are then concatenated with text tokens, $\mathcal{Q} = \{\sigma_N, \sigma_{N+1}, \dots, \sigma_{N+M-1}\}$, forming a unified input sequence of length $N+M$.

\paragraph{Next token prediction via LLM.}
The concatenated token sequence is processed by LVLM (parametrized by $\theta$) in an auto-regressive manner.
The model computes logits $\ell_t$ for each potential next token:
\begin{equation}
\ell_t = \log p(\xi_t | \mathcal{V}, \mathcal{Q}, \xi_{<t}; \theta),
\end{equation}
where $\xi_t$ is the token predicted at timestep $t$, and $\xi_{<t}$ denotes the sequence generated up to timestep $(t - 1)$.
These logits are converted into a probability distribution via the softmax function:
\begin{equation}
p(\xi_t) =  \mathrm{Softmax}(\ell_t).
\end{equation}
from which the next token is sampled.

\vspace{\subsecmargin}\subsection{Attentional Vision Calibration}

\begin{figure}[t!]
    \begin{minipage}[t!]{0.49\linewidth}
    \centering
    \vspace{-1mm}
    \includegraphics[width=\linewidth]{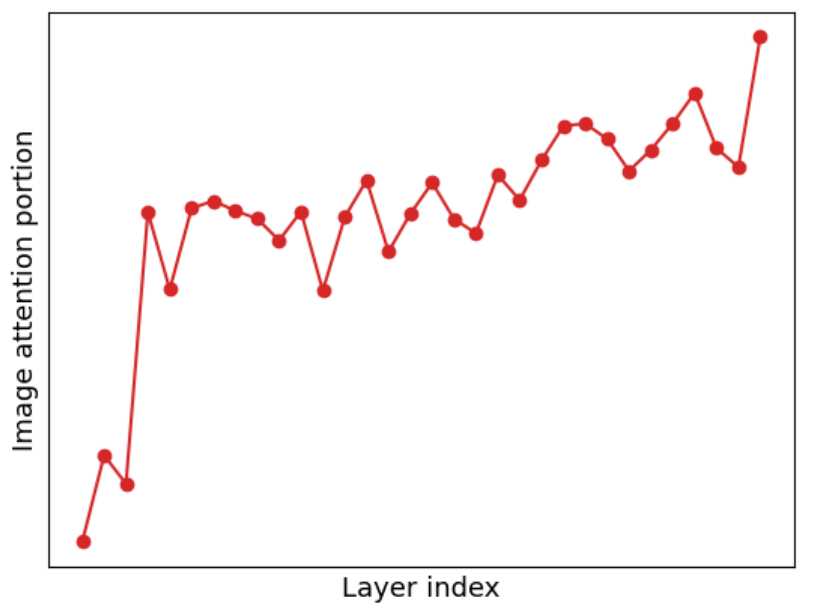}\\
    {\footnotesize (a) InstructBLIP}
    \end{minipage}
    \hfill
    \begin{minipage}[t!]{0.49\linewidth}
    \centering
    \vspace{-1mm}
    \includegraphics[width=\linewidth]{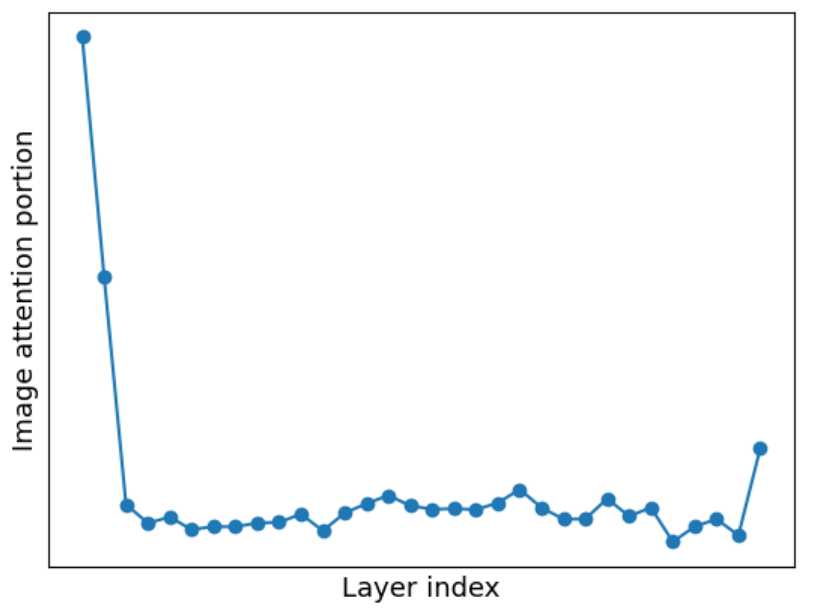}\\
    {\footnotesize (b) LLaVA-1.5}
    \end{minipage}
    \vspace{-2mm}
    \caption{
    \textbf{Layer-wise image attention proportion in LVLMs~\citep{liu2023improved,dai2024instructblip}.}
    This shows the proportion of attention given to image tokens at each layer relative to total attention.
    Different layers exhibit distinct attention patterns, which vary across models.
    Attention weights are averaged over 60 questions from the LLaVA-Bench~\citep{liu2023visual}.
    }
    \label{fig:layer_attention}
\end{figure}
Visual hallucinations in LVLMs often arise during decoding when the model’s token selection is guided by skewed probability distributions that do not faithfully reflect the underlying visual input. Our observations (see~\cref{sec:observations}) indicate that this problem is linked to an attentional bias toward certain non-relevant tokens, which we term \emph{blind tokens}. Our approach aims to recalibrate these attention patterns to mitigate hallucinations.

\paragraph{Layer selection.}
Different layers in LVLMs contribute variably to processing visual information.
As illustrated in \cref{fig:layer_attention}, models such as InstructBLIP~\citep{dai2024instructblip} and LLaVA-1.5~\citep{liu2023visual} exhibit different attention distributions across layers. To accommodate these differences, we first select layers that exhibit a high proportion of attention on image tokens.
%
Formally, for the $i$th layer, we define the attention weight matrix:
\begin{equation}
    \mathbf{A}_i = \left[ \mathbf{a}_{h,q,k}^i \right]_{(h,q,k)=(1,1,1)}^{(h,q,k)=(H,N+M,N+M)},
\end{equation}
where $\mathbf{a}_{h,q,k}^i$ represents the attention weight from head $h$, for query $q$ to key $l_k$ in layer $i$.
With image tokens $\mathcal{V} \in \mathbb{R}^{N \times D}$ and query tokens $\mathcal{Q} \in \mathbb{R}^{M \times D}$, we compute the proportion of attention dedicated to image tokens in layer $i$:
%
\begin{equation}
    AP_i^{\text{layer}} =\frac{\sum_{h}\sum_{k=1}^{N} \mathbf{a}^i_{h, (N+M), k}}
    {\sum_{i, h}\sum_{k=1}^{N} \mathbf{a}^i_{h, (N+M), k}},
\end{equation}
where $H$ is the total number of attention heads, $N$ is the number of image tokens, and $M$ is the number of query tokens.
We then sort the layers by this proportion and select layers using top-P sampling with threshold $\gamma$:
\begin{equation}
    \{\text{Selected Layers}\} = \text{top-P}(\{AP_i^{\text{layer}}\}_{i=1}^{L}, \gamma).
\label{eq:layer_selection}
\end{equation}
These selected layers provide the basis for further token-level analysis.

\paragraph{Blind token identification.}
Within the selected layers, we compute the attention proportion for each image token by averaging over all heads:
\begin{equation}
    AP^{\text{image}} = \frac{\sum_{i \in \{\text{Selected Layers}\}} \sum_{h=1}^{H}
    \mathbf{a}^i_{h, (N+M), [1:N]}}{| \{\text{Selected Layers}\} | \times H }. 
\end{equation}
%
To identify tokens that disproportionately capture attention, \ie, \textit{blind tokens}, we calculate the mean ($\mu$) and standard deviation ($\sigma$) of the image attention weights.
Tokens with an attention proportion exceeding $\mu + \lambda\sigma$ (where $\lambda$ is a hyperparameter) are classified as blind tokens:
\begin{equation}
    \{ \text{Blind Token Indices} \} = \{ j | AP^{\text{image}}_j > \mu + \lambda\sigma \}.
\label{eq:blind_token_identification}
\end{equation}

\paragraph{Contrastive decoding.}
To mitigate hallucinations, we reduce the influence of blind tokens during decoding via contrastive decoding~\citep{leng2023mitigating,favero2024multi}. We construct a biased set of visual tokens by zeroing out non-blind tokens:
\begin{equation}
    \mathcal{V}^{\ast} = \bigcup_{j=1}^{N} \vmathbb{1}_{\{j \in \text{Blind Token Indices}\}}(j) \nu_j.
\end{equation}
We then compute the logits for the next token using both the original visual tokens ($\mathcal{V}$) and the biased tokens ($\mathcal{V}^{\ast}$):
\begin{equation}
\begin{aligned}
    \ell_t &= \log p(\xi_t | \mathcal{V}, \mathcal{Q}, \xi_{<t}; \theta), \\
    \ell_t^{\ast} &= \log p(\xi_t | \mathcal{V^{\ast}}, \mathcal{Q}, \xi_{<t}; \theta),
\end{aligned}
\end{equation}
%
Finally, we adjust the logits using a contrastive scheme and sample the next token from:
\begin{equation}
    \xi_t \sim \mathrm{Softmax}((1+\alpha)\ell_t - \alpha \ell_t^{\ast}).
\label{eq:contrastive_decoding}
\end{equation}
Here, $\alpha$ controls the strength of the contrast. This adjustment effectively down-weights the contribution of blind tokens and promotes a more balanced attention distribution, thereby reducing hallucinations in the final output.

\begin{table*}[t]
    
    \centering
    \small
    \setlength\tabcolsep{5pt} 
    \scalebox{0.8}{
    \begin{tabular}{lllx{36}x{36}x{36}x{36}x{36}x{36}x{36}x{36}}
    \toprule
     & \multirow{2}{*}{\textbf{Setup}} & \multirow{2}{*}{\textbf{Method}} & \multicolumn{4}{c}{\textbf{InstructBLIP~\citep{dai2024instructblip}}} & \multicolumn{4}{c}{\textbf{LLaVA-1.5~\citep{liu2023visual}}}\\
    \arrayrulecolor{gray} \cmidrule(lr){4-7} \cmidrule(lr){8-11}
     &  &  & {{Acc.\up}} & {{Prec.\up}} & {{Rec.\up}} & {{F1\up}} & {{Acc.\up}} & {{Prec.\up}} & {{Rec.\up}} & {{F1\up}} \\
    \midrule
    \multirow{15}{*}{\rot{\textbf{\normalsize MS-COCO}\quad}} & \multirow{4}{*}{Random} & \textit{base} & 82.27 & 82.84 & 81.40 & 82.11 & 84.47 & 83.35 & 86.13 & 84.72 \\
     &  & VCD & 83.37 & 83.39 & 82.60 & 83.24 & 84.80 & 83.00 & \colorbox{Best}{87.53} & 85.20 \\
     &  & M3ID & \colorbox{SecondBest}{84.37} & \colorbox{SecondBest}{84.62} & \colorbox{Best}{84.00} & \colorbox{SecondBest}{84.31} & \colorbox{SecondBest}{86.00} & \colorbox{SecondBest}{85.11} & \colorbox{SecondBest}{87.27} & \colorbox{SecondBest}{86.18} \\
     &  & \textbf{\Ours} & \colorbox{Best}{88.73} & \colorbox{Best}{93.88} & \colorbox{SecondBest}{82.87} & \colorbox{Best}{88.03} & \colorbox{Best}{87.93} & \colorbox{Best}{88.24} & \colorbox{Best}{87.53} & \colorbox{Best}{87.88} \\
     \arrayrulecolor{gray!50}\cmidrule(lr){3-11}
      & \multirow{4}{*}{Popular} & \textit{base} &  77.77 &  74.81   & 83.73 & 79.02 & 82.23   & \colorbox{SecondBest}{79.72} &    86.47 &  82.95 \\
     &  & VCD & \colorbox{SecondBest}{78.00} &   \colorbox{SecondBest}{75.12} &  83.73 &  \colorbox{SecondBest}{79.19} & 82.27 & 79.19 &  87.53   &83.15  \\
     &  & M3ID & 77.30 & 74.10 & \colorbox{SecondBest}{83.93} & 78.71 & \colorbox{SecondBest}{82.83} & 79.62 & \colorbox{SecondBest}{88.27} & \colorbox{SecondBest}{83.72} \\
     &  & \textbf{\Ours} & \colorbox{Best}{83.90} &  \colorbox{Best}{81.33} &   \colorbox{Best}{88.00} & \colorbox{Best}{84.53} & \colorbox{Best}{84.33} &   \colorbox{Best}{81.71} &   \colorbox{Best}{88.47} & \colorbox{Best}{84.96} \\
     \arrayrulecolor{gray!50}\cmidrule(lr){3-11} 
      & \multirow{4}{*}{Adversarial} & \textit{base} & 73.13 &  69.41   & 82.60 &   75.46 & 77.10 & 72.57   & 87.13 &   79.19   \\
     &  & VCD & 75.87   & \colorbox{SecondBest}{72.85}  & 82.47  & 77.36 & 76.10  & 71.50 & 86.80 & 78.41  \\
     &  & M3ID & \colorbox{SecondBest}{76.03} & 72.47 & \colorbox{Best}{83.93} & \colorbox{SecondBest}{77.79} & \colorbox{Best}{77.70} & \colorbox{Best}{73.23} & \colorbox{SecondBest}{87.33} & \colorbox{Best}{79.66} \\
     &  & \textbf{\Ours} & \colorbox{Best}{81.57}    & \colorbox{Best}{80.37}    &\colorbox{SecondBest}{83.53}    &\colorbox{Best}{81.92} & \colorbox{SecondBest}{77.53}  &\colorbox{SecondBest}{72.82}    &\colorbox{Best}{87.87}    &\colorbox{SecondBest}{79.64}    \\
     \arrayrulecolor{gray}\midrule
     \multirow{15}{*}{\rot{\textbf{\normalsize A-OKVQA}}} & \multirow{4}{*}{Random} & \textit{base} & 81.00 &    77.71   & 86.93 & 82.06 & 82.73 &  77.43 &  92.40 &  84.26 \\
     &  & VCD & 81.73   & \colorbox{SecondBest}{78.67} & 87.07 &  82.66 & 81.30   & 75.45  & 92.80 & 83.23  \\
     &  & M3ID & \colorbox{SecondBest}{82.33} & 77.81 & \colorbox{Best}{90.47} & \colorbox{SecondBest}{83.66} & \colorbox{SecondBest}{83.57} & \colorbox{SecondBest}{77.86} & \colorbox{Best}{93.80} & \colorbox{SecondBest}{85.09} \\
     &  & \textbf{\Ours} & \colorbox{Best}{88.47}    &\colorbox{Best}{87.66}    &\colorbox{SecondBest}{89.53}    &\colorbox{Best}{88.59} & \colorbox{Best}{84.60}  &\colorbox{Best}{79.29}    &\colorbox{SecondBest}{93.67}    &\colorbox{Best}{85.88}    \\
     \arrayrulecolor{gray!50}\cmidrule(lr){3-11} 
      & \multirow{4}{*}{Popular} & \textit{base} & 75.00 &   70.14 &  87.07 &  77.69 & 76.10   & 69.86 & 91.80 &  79.34 \\
     &  & VCD & 75.33   & \colorbox{SecondBest}{70.52}  & 87.07  & 77.92 & {75.43} &  {68.58} &    \colorbox{SecondBest}{93.87} &  79.26   \\
     &  & M3ID & \colorbox{SecondBest}{75.60} & 70.40 & \colorbox{SecondBest}{88.33} & \colorbox{SecondBest}{78.36} & \colorbox{SecondBest}{76.80} & \colorbox{SecondBest}{70.20} & 93.13 & \colorbox{SecondBest}{80.06} \\
     &  & \textbf{\Ours} & \colorbox{Best}{81.77}    &\colorbox{Best}{77.82}    &\colorbox{Best}{88.87}    &\colorbox{Best}{82.98} & \colorbox{Best}{78.83}  & \colorbox{Best}{72.10}   &\colorbox{Best}{94.07}    &\colorbox{Best}{81.63}    \\
     \arrayrulecolor{gray!50}\cmidrule(lr){3-11}
      & \multirow{4}{*}{Adversarial} & \textit{base} & 68.80 &   63.57 & 88.07 & 73.84 & 67.90 & \colorbox{SecondBest}{62.11} & 91.80 & 74.09 \\
     &  & VCD & \colorbox{SecondBest}{69.70} &   \colorbox{SecondBest}{64.54} &   87.47 &  74.27 & {67.43} &   {61.50}&    93.20 &  74.11 \\
     &  & M3ID & 69.57 & 64.21 & \colorbox{Best}{88.40} & \colorbox{SecondBest}{74.39} & \colorbox{SecondBest}{68.10} & 61.99 & \colorbox{SecondBest}{93.60} & \colorbox{SecondBest}{74.58} \\
     &  & \textbf{\Ours} & \colorbox{Best}{72.53}    &\colorbox{Best}{67.12}    &\colorbox{SecondBest}{88.33}    &\colorbox{Best}{76.28} & \colorbox{Best}{68.97}& \colorbox{Best}{62.70} &\colorbox{Best}{93.67}&   \colorbox{Best}{75.11} \\
     \arrayrulecolor{gray}\midrule
     \multirow{15}{*}{\rot{\textbf{\normalsize GQA}}} & \multirow{4}{*}{Random} & \textit{base} & 80.00 &   77.08   & 85.40 & 81.02 & 82.40   & \colorbox{SecondBest}{77.03}  & 92.33  & 83.99 \\
     &  & VCD & \colorbox{SecondBest}{81.73} &  \colorbox{SecondBest}{79.35} &  {85.80} & \colorbox{SecondBest}{82.45} & 82.27  & 75.85  & \colorbox{SecondBest}{94.67}  & 84.22 \\
     &  & M3ID & 80.57 & 76.77 & \colorbox{Best}{87.67} & 81.85 & \colorbox{SecondBest}{82.83} & 76.64 & 94.47 & \colorbox{SecondBest}{84.62} \\
     &  & \textbf{\Ours} & \colorbox{Best}{86.47}    &\colorbox{Best}{85.89} & \colorbox{SecondBest}{87.27} &\colorbox{Best}{86.57} & \colorbox{Best}{85.00}  &\colorbox{Best}{78.81}    &\colorbox{Best}{95.73}    &\colorbox{Best}{86.45}    \\
     \arrayrulecolor{gray!50}\cmidrule(lr){3-11}
      & \multirow{4}{*}{Popular} & \textit{base} & 73.53 &  68.80 &     86.13 &     76.49 & 72.03 &     65.57 &     92.80 &     76.84  \\
     &  & VCD & {74.10} &   \colorbox{SecondBest}{69.45}&    {86.07}&    {76.87} & {71.77}   & {64.90}&  \colorbox{SecondBest}{94.80} &  77.05 \\
     &  & M3ID & \colorbox{SecondBest}{74.57} & \colorbox{SecondBest}{69.45} & \colorbox{Best}{87.83} & \colorbox{SecondBest}{77.53} & \colorbox{SecondBest}{72.83} & \colorbox{SecondBest}{66.04} & 94.00 & \colorbox{SecondBest}{77.58} \\
     &  & \textbf{\Ours} & \colorbox{Best}{78.00}    & \colorbox{Best}{73.68}   & \colorbox{SecondBest}{87.13}   & \colorbox{Best}{79.84} & \colorbox{Best}{74.80} &   \colorbox{Best}{67.46}&    \colorbox{Best}{95.80} &   \colorbox{Best}{79.17} \\
     \arrayrulecolor{gray!50}\cmidrule(lr){3-11}
      & \multirow{4}{*}{Adversarial} & \textit{base} & 68.00    & 63.49 & 84.73 & 72.59 & \colorbox{SecondBest}{68.73} &    \colorbox{SecondBest}{62.54} &  93.40   & \colorbox{SecondBest}{74.92} \\
     &  & VCD & \colorbox{SecondBest}{70.27} &   \colorbox{SecondBest}{65.43} & {85.93} & \colorbox{SecondBest}{74.29} & 68.27 & {62.00}   & 94.40 & 74.84 \\
     &  & M3ID & 68.90 & 64.06 & \colorbox{SecondBest}{86.13} & 73.47 & 68.13 & 61.88 & \colorbox{SecondBest}{94.47} & 74.78 \\
     &  & \textbf{\Ours} & \colorbox{Best}{73.07}    & \colorbox{Best}{67.80}&  \colorbox{Best}{87.87}&    \colorbox{Best}{76.54} & \colorbox{Best}{69.20}   & \colorbox{Best}{62.61}   & \colorbox{Best}{95.33}&  \colorbox{Best}{75.58} \\
    \bottomrule
    \end{tabular}
    }
    \vspace{\abovetabcapmargin}
    \caption{
        \textbf{POPE benchmark results.}
        \Ours consistently outperforms \textit{base} decoding and other methods: VCD~\citep{leng2023mitigating} and M3ID~\citep{favero2024multi}.
        We reimplemented VCD and M3ID in our evaluation setup.
    }
    \label{tab:POPE}
\end{table*}

\vspace{\secmargin}\section{Experiments}\vspace{\secmargin}
\label{sec:experiments}
Additional experimental results are provided in~\cref{sec:apependix_additional_experiments}.

\subsection{Evaluation Setup}
We deliberately avoid constraining LVLMs to provide one-word responses (\eg, “Yes” or “No”) for discriminative tasks, as our analysis (see~\cref{tab:POPE_one_word}) shows that such constraints can significantly alter performance.
For our experiments, we set $\text{P}=0.5$ in~\cref{eq:layer_selection}, $\lambda=1$ in~\cref{eq:blind_token_identification}, and $\alpha=3$ for InsturctBLIP~\citep{dai2024instructblip} and $\alpha=2.5$ for LLaVA-1.5~\citep{liu2023visual} in~\cref{eq:contrastive_decoding}.
\footnote{Further experimental details are provided in~\cref{sec:appendix_experimental_details}.}


\paragraph{LVLMs.}
We evaluate \Ours on two state-of-the-art LVLMs: \textbf{InstructBLIP}\citep{dai2024instructblip} and \textbf{LLaVA-1.5}\citep{liu2023visual}, both of which use Vicuna 7B~\citep{chiang2023vicuna} as the LLM backbone. InstructBLIP employs the Q-Former~\citep{li2023blip} to efficiently fuse visual and textual features using a fixed number of tokens (\eg, 32 tokens), while LLaVA-1.5 aligns image and text modalities via linear projection layers. Notably, \Ours is model-agnostic and can be integrated into various LVLM architectures.


\paragraph{Benchmarks.}
\textbf{(1) POPE}~\citep{li2023evaluating} views hallucination evaluation as a binary classification task (yes/no) with questions on object presence (\eg, "Is there a cat in the image?").
It evaluates both visible and imaginary objects across across three setups: random, popular, and adversarial.
\textbf{(2) MME}~\citep{fu2024mme} evaluates 14 subtasks---including object existence, count, position, and color---via binary questions.
\textbf{(3) AMBER}~\citep{wang2023llm}combines generative and discriminative tasks, focusing on hallucinations in object existence, attributes, and relationships. Generative performance is measured by CHAIR, discriminative by F1, with the overall AMBER score computed as $((100 - \text{CHAIR}) + \text{F1}) / 2$.


\paragraph{Baselines.}
We compare \Ours against recent contrastive decoding methods, notably \textbf{VCD}\citep{leng2023mitigating} and \textbf{M3ID}\citep{favero2024multi}, which mitigate hallucinations by enhancing the reference image’s influence relative to the language model’s prior through contrasting output distributions from original and altered visual inputs. Our reimplementations of VCD and M3ID serve as baselines, as they too avoid external models, costly self-feedback, and additional training.

\begin{table*}[t!]
    \vspace{-1mm}
    \centering
    \small
    \setlength\tabcolsep{4pt} 
    \scalebox{0.85}{
    \begin{tabular}{llx{55}x{55}x{55}x{55}x{55}}
        \toprule
        \multirow{2}{*}{\textbf{Model}} & \multirow{2}{*}{\textbf{Method}} & \multicolumn{2}{c}{\textbf{Object-level}} & \multicolumn{2}{c}{\textbf{Attribute-level}} & \multicolumn{1}{c}{\multirow{2}{*}{\textbf{\makecell{Total \\ Score}}}}\\
        \arrayrulecolor{gray} \cmidrule(lr){3-4} \cmidrule(lr){5-6} 
         & & Existence\up & Count\up & Position\up & Color\up & \\
        \arrayrulecolor{gray} \midrule
         \multirow{5}{*}{\textbf{InstructBLIP}} & \textit{base} & 170.19$_{(\pm 11.12)}$ & 89.52$_{(\pm 11.04)}$ & 67.62$_{(\pm 14.04)}$ & 114.76$_{(\pm 9.60)}$ & 442.09$_{(\pm 31.51)}$ \\
         & VCD & 172.62$_{(\pm 8.92)}$ & \colorbox{Best}{98.33$_{(\pm 15.99)}$} & 71.90$_{(\pm 13.42)}$ & \colorbox{SecondBest}{117.14$_{(\pm 10.70)}$} & \colorbox{SecondBest}{459.99$_{(\pm 16.56)}$} \\
         & M3ID & \colorbox{SecondBest}{173.89$_{(\pm 10.52)}$} & \colorbox{SecondBest}{89.72$_{(\pm 13.44)}$} & \colorbox{SecondBest}{72.72$_{(\pm 14.77)}$} & 110.56$_{(\pm 7.20)}$ & 446.88$_{(\pm 28.54)}$ \\
         & \textbf{\Ours} & \colorbox{Best}{184.76$_{(\pm 5.56)}$} & 82.85$_{(\pm 12.16)}$ & \colorbox{Best}{74.76$_{(\pm 6.19)}$} & \colorbox{Best}{131.43$_{(\pm 4.76)}$} & \colorbox{Best}{473.80$_{(\pm 19.67)}$} \\
        \arrayrulecolor{gray} \midrule
        \multirow{5}{*}{\textbf{LLaVA 1.5}} & \textit{base} & 173.57$_{(\pm 8.16)}$ & \colorbox{SecondBest}{110.00$_{(\pm 15.82)}$} & 100.47$_{(\pm 18.78)}$ & 125.24$_{(\pm 15.91)}$ & 509.28$_{(\pm 30.57)}$ \\
         & VCD & 172.14$_{(\pm 8.09)}$ & \colorbox{Best}{117.14$_{(\pm 8.76)}$} & \colorbox{SecondBest}{103.33$_{(\pm 20.56)}$} & 119.52$_{(\pm 8.58)}$ & \colorbox{SecondBest}{512.14$_{(\pm 31.82)}$} \\
         & M3ID & \colorbox{SecondBest}{178.33$_{(\pm 6.83)}$} & 107.22$_{(\pm 14.78)}$ & 96.39$_{(\pm 5.52)}$ & \colorbox{SecondBest}{127.50$_{(\pm 8.28)}$} & 509.44$_{(\pm 22.52)}$ \\
         & \textbf{\Ours} & \colorbox{Best}{189.29$_{(\pm 1.82)}$} & 104.76$_{(\pm 11.66)}$ & \colorbox{Best}{106.19$_{(\pm 13.93)}$} & \colorbox{Best}{127.86$_{(\pm 9.13)}$} & \colorbox{Best}{528.09$_{(\pm 24.70)}$} \\
        \bottomrule
    \end{tabular}
    }
    \vspace{\abovetabcapmargin}
    \caption{
        \textbf{MME-Hallucination~\citep{fu2024mme} benchmark results.}
        Our method effectively reduces hallucinations at both object and attribute levels, surpassing VCD~\citep{leng2023mitigating} and M3ID~\citep{favero2024multi} in Total Score.
    }
    \vspace{-1mm}
    \label{tab:MME}
\end{table*}
\begin{figure*}[t!]
    \begin{minipage}[t!]{0.49\linewidth}
    \centering
    \includegraphics[width=\textwidth]{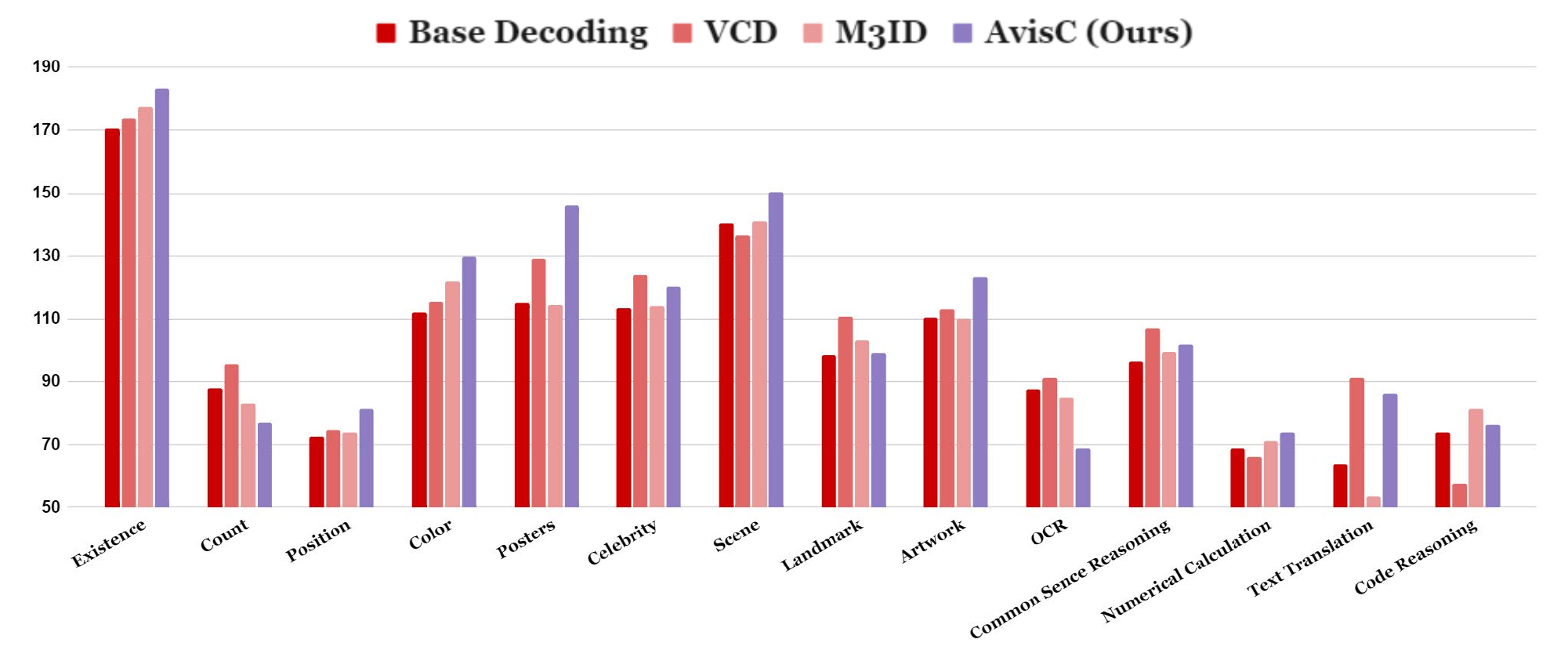}\\
    \vspace{-3mm}
    {\footnotesize (a) InstructBLIP~\citep{dai2024instructblip}}
    \end{minipage}
    \hfill
    \begin{minipage}[t!]{0.49\linewidth}
    \centering
    \includegraphics[width=\textwidth]{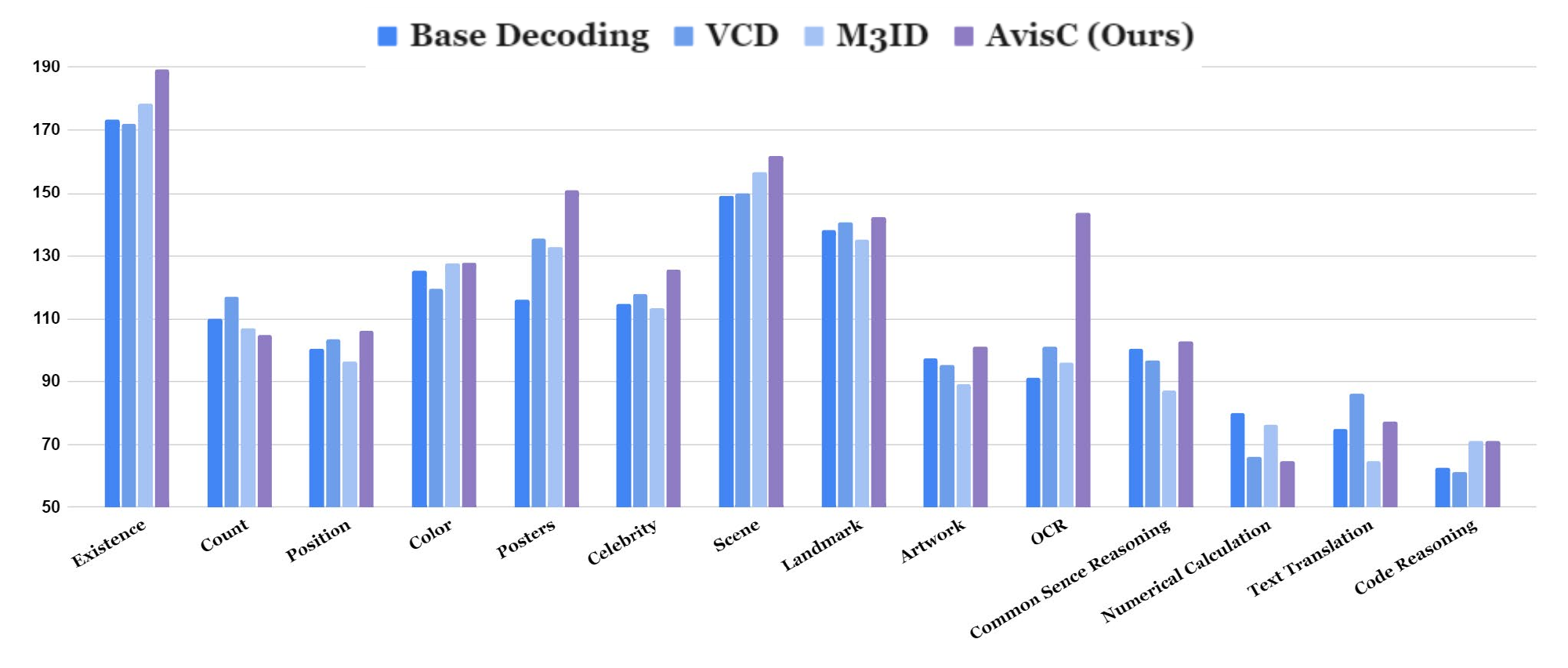}\\
    \vspace{-3mm}
    
    {\footnotesize (b) LLaVA-1.5~\citep{liu2023visual}}
    \end{minipage}
    \vspace{-2mm}
    \caption{
    \textbf{Performance comparison on MME-Fullset.}
    \Ours achieves top performance in 7 of 14 categories with InstructBLIP~\citep{dai2024instructblip} and in 11 categories with LLaVA-1.5~\citep{liu2023visual}.
    Beyond minimizing hallucinations, \Ours also boosts the general functionality of LVLMs.
    }%
    \label{fig:mme_full}
\end{figure*}
\begin{table*}[t]
    
    \centering
    \small
    \renewcommand{\arraystretch}{1.45} 
    \setlength\tabcolsep{4pt} 
    \scalebox{0.8}{
    \begin{tabular}{lrx{45}x{45}x{45}x{45}x{0}x{45}x{45}x{45}x{45}}
    \toprule
     & \multirow{2}{*}{\textbf{Metric}} & \multicolumn{4}{c}{\textbf{InstructBLIP~\citep{dai2024instructblip}}} &  & \multicolumn{4}{c}{\textbf{LLaVA 1.5~\citep{liu2023visual}}}\\
    \arrayrulecolor{gray} \cmidrule(lr){3-6} \cmidrule(lr){8-11}
     &  & {\textit{base}} & {{VCD}} & {{M3ID}} & {\textbf{\Ours}} & & {\textit{base}} & {{VCD}} & {{M3ID}} & {\textbf{\Ours}} \\
    \cmidrule(lr){1-2} \cmidrule(lr){3-6} \cmidrule(lr){8-11}
    \multirow{4}{*}{\rot{\textbf{Generative}}}
    & CHAIR{\down}
    & {8.40$_{(\pm 0.57)}$} & {7.60$_{(\pm 0.42)}$} & \colorbox{SecondBest}{6.85$_{(\pm 0.07)}$} & \colorbox{Best}{6.70$_{(\pm 0.28)}$} & 
    & {7.95$_{(\pm 0.64)}$} & {6.70$_{(\pm 0.42)}$} & \colorbox{Best}{6.00$_{(\pm 0.14)}$} & \colorbox{SecondBest}{6.25$_{(\pm 0.07)}$} \\
     & Cover{\up}
     & {46.40$_{(\pm 1.27)}$} & \colorbox{Best}{47.65$_{(\pm 0.35)}$} & \colorbox{SecondBest}{47.20$_{(\pm 0.71)}$} & {46.65$_{(\pm 1.48)}$} & 
     & {44.45$_{(\pm 0.21)}$} & {46.50$_{(\pm 0.28)}$} & \colorbox{Best}{48.90$_{(\pm 0.28)}$} & \colorbox{SecondBest}{46.55$_{(\pm 0.64)}$} \\
     & Hal{\down}
     & {31.10$_{(\pm 0.64)}$} & {29.90$_{(\pm 0.99)}$} & \colorbox{Best}{27.50$_{(\pm 0.71)}$} & \colorbox{SecondBest}{28.00$_{(\pm 0.28)}$} & 
     & {31.00$_{(\pm 2.83)}$} & {27.80$_{(\pm 1.70)}$} & \colorbox{SecondBest}{26.00$_{(\pm 0.28)}$} & \colorbox{Best}{25.60$_{(\pm 1.70)}$} \\
     & Cog{\down}
     & {2.60$_{(\pm 0.05)}$} & \colorbox{Best}{2.20$_{(\pm 0.14)}$} & \colorbox{Best}{2.20$_{(\pm 0.14)}$} & \colorbox{SecondBest}{2.55$_{(\pm 0.35)}$} & 
     & {2.15$_{(\pm 0.35)}$} & \colorbox{SecondBest}{1.95$_{(\pm 0.35)}$} & \colorbox{Best}{1.45$_{(\pm 0.07)}$} & {2.00$_{(\pm 0.04)}$} \\
    \arrayrulecolor{gray} \cmidrule(lr){1-2} \cmidrule(lr){3-6} \cmidrule(lr){8-11}
    \multirow{4}{*}{\rot{\textbf{Discriminative}}}
    & Acc.{\up}
    & {68.20$_{(\pm 0.14)}$} & {69.65$_{(\pm 0.35)}$} & \colorbox{SecondBest}{69.05$_{(\pm 0.35)}$} & \colorbox{Best}{72.60$_{(\pm 0.42)}$} & 
    & {67.00$_{(\pm 0.71)}$} & \colorbox{SecondBest}{67.30$_{(\pm 1.41)}$} & {67.25$_{(\pm 0.21)}$} & \colorbox{Best}{70.70$_{(\pm 0.57)}$} \\
     & Prec.{\up}
     & {79.00$_{(\pm 0.14)}$} & \colorbox{Best}{80.70$_{(\pm 0.42)}$} & \colorbox{SecondBest}{79.70$_{(\pm 0.28)}$} & {72.60$_{(\pm 0.42)}$} & 
     & {85.45$_{(\pm 0.49)}$} & \colorbox{SecondBest}{86.10$_{(\pm 1.70)}$} & \colorbox{Best}{86.50$_{(\pm 0.57)}$} & {85.45$_{(\pm 0.21)}$} \\
     & Rec.{\up}
     & {70.70$_{(\pm 0.42)}$} & \colorbox{SecondBest}{71.60$_{(\pm 0.42)}$} & {71.25$_{(\pm 0.35)}$} & \colorbox{Best}{76.10$_{(\pm 0.05)}$} & 
     & \colorbox{SecondBest}{60.95$_{(\pm 1.20)}$} & {60.55$_{(\pm 1.34)}$} & {60.05$_{(\pm 0.07)}$} & \colorbox{Best}{67.55$_{(\pm 0.92)}$} \\
     & F1{\up}
     & {74.60$_{(\pm 0.14)}$} & \colorbox{SecondBest}{75.90$_{(\pm 0.42)}$} & {75.25$_{(\pm 0.07)}$} & \colorbox{Best}{78.60$_{(\pm 0.28)}$} & 
     & \colorbox{SecondBest}{71.10$_{(\pm 0.99)}$} & \colorbox{SecondBest}{71.10$_{(\pm 1.56)}$} & {70.90$_{(\pm 0.14)}$} & \colorbox{Best}{75.45$_{(\pm 0.64)}$} \\
    \arrayrulecolor{gray} \cmidrule(lr){1-2} \cmidrule(lr){3-6} \cmidrule(lr){8-11}
    \multicolumn{2}{r}{\textbf{AMBER{\up}}}
    & {83.10$_{(\pm 0.35)}$} & {84.15$_{(\pm 0.05)}$} & \colorbox{SecondBest}{84.20$_{(\pm 0.07)}$} & \colorbox{Best}{85.95$_{(\pm 0.05)}$} & 
    & {81.58$_{(\pm 0.18)}$} & {82.20$_{(\pm 0.99)}$} & \colorbox{SecondBest}{82.45$_{(\pm 0.14)}$} & \colorbox{Best}{84.60$_{(\pm 0.35)}$} \\
    \bottomrule
    \end{tabular}
    }
    \vspace{\abovetabcapmargin}
    \caption{
        \textbf{AMBER~\citep{wang2023llm} benchmark results.}
        \Ours outperforms contrastive decoding baselines~\citep{leng2023mitigating,favero2024multi}
        in both generative and discriminative tasks, achieving the highest AMBER score.
    }
    \label{tab:AMBER}
\end{table*}

\vspace{\subsecmargin}\subsection{Benchmark Results}

\paragraph{POPE.}
\Cref{tab:POPE} summarizes performance on the POPE benchmark~\citep{li2023evaluating} across MS-COCO~\citep{lin2014microsoft}, A-OKVQA~\citep{schwenk2022okvqa}, and GQA~\citep{hudson2019gqa} datasets under the Random, Popular, and Adversarial setups. Overall, \Ours consistently outperforms the baseline (\textit{base}) and other decoding methods (VCD, M3ID) in most cases, achieving the highest Accuracy and F1 scores. Additionally, balanced improvements in Precision and Recall suggest reduced errors and better information capture. For InstructBLIP, \Ours yields a significant performance boost---especially in mitigating hallucinations related to object existence---while LLaVA-1.5 shows somewhat lower gains in the more challenging Popular and Adversarial setups. Nonetheless, \Ours proves robust across different datasets and query configurations.







\paragraph{MME-Hallucination.}
\Cref{tab:MME} presents performance results for InstructBLIP and LLaVA-1.5 on the MME-Hallucination benchmark~\citep{fu2024mme}. We evaluate both object-level metrics (Existence, Count) and attribute-level metrics (Position, Color). Both models show marked improvements in the Existence category when using \Ours, achieving the highest scores. While VCD slightly outperforms in the Count metric, \Ours excels in Position and Color, leading to superior Total Scores overall. These results affirm that \Ours effectively reduces hallucinations and improves accuracy across multiple dimensions.





\paragraph{MME-Fullset.}
\Cref{fig:mme_full} compares various decoding methods on the MME-Fullset benchmark~\citep{fu2024mme} across 14 categories. \Ours achieves top results in 7 categories for InstructBLIP and 11 for LLaVA-1.5. This indicates that \Ours enhances the model’s ability to extract and utilize informative visual features through attention calibration. Although both models experience a slight decline in the Count category with \Ours---and InstructBLIP shows lower performance on OCR tasks---LLaVA-1.5 sees significant OCR improvements, demonstrating that the impact of \Ours can vary across different models. Overall, \Ours delivers superior results across most tasks compared to the baselines.






\paragraph{AMBER.}
\Cref{tab:AMBER} shows results on the AMBER benchmark~\citep{wang2023llm}, which includes both generative and discriminative tasks. 
\Ours significantly improves discriminative performance (Accuracy and F1) for both InstructBLIP and LLaVA-1.5, outperforming Base, VCD, and M3ID.
In generative tasks, it also achieves substantial gains, particularly in the Existence metric, indicating better object detection. Overall, \Ours enables both models to achieve the highest scores across most AMBER metrics.










\begin{table*}[t!]
    
    \begin{minipage}[t!]{0.32\textwidth}
        \begin{center}
        \begin{small}
        \setlength\tabcolsep{3pt} 
        \scalebox{0.74}{
        \begin{tabular}{lx{25}x{25}x{25}x{25}c}
            \toprule
             \multicolumn{6}{c}{\textbf{\normalsize (a) InstructBLIP~\citep{dai2024instructblip} ($\lambda=1$)}} \\
             \arrayrulecolor{gray} \midrule
             {} & \multicolumn{2}{c}{\textbf{Object}} & \multicolumn{2}{c}{\textbf{Attribute}} & \multicolumn{1}{c}{\multirow{2}{*}{\textbf{\makecell{Total \\ Score}}}} \\
             \arrayrulecolor{gray} \cmidrule(lr){2-3} \cmidrule(lr){4-5} 
             $\alpha$ & Exist. & Count & Position & Color & {} \\
             \arrayrulecolor{gray} \midrule
             0.5 & {180} & {83.33} & {80.00} & {130} & {473.33} \\
             2.0 & {180} & {86.66} & {75.00} & {135} & {476.66} \\
             2.5 & {180} & {85.00} & {71.66} & {135} & {471.66} \\
             \textbf{3.0} & \textbf{195} & \textbf{75.00} & \textbf{73.33} & \textbf{135} & \textbf{478.33} \\
            \bottomrule
        \end{tabular}
        }
        \end{small}
        \end{center}
    \end{minipage}
    \hfill
    \begin{minipage}[t!]{0.32\textwidth}
        \begin{center}
        \begin{small}
        \setlength\tabcolsep{3pt} 
        \scalebox{0.74}{
        \begin{tabular}{lx{25}x{25}x{25}x{25}c}
            \toprule
             \multicolumn{6}{c}{\textbf{\normalsize (b) InstructBLIP~\citep{dai2024instructblip} ($\alpha=3$)}} \\
             \arrayrulecolor{gray} \midrule
             & \multicolumn{2}{c}{\textbf{Object}} & \multicolumn{2}{c}{\textbf{Attribute}} & \multicolumn{1}{c}{\multirow{2}{*}{\textbf{\makecell{Total \\ Score}}}} \\
             \arrayrulecolor{gray} \cmidrule(lr){2-3} \cmidrule(lr){4-5} 
             $\lambda$ & Exist. & Count & Position & Color & \\
             \arrayrulecolor{gray} \midrule
             0.0 & {180} & {75.00} & {60.00} & {115.00} & {430.00} \\
             0.1 & {185} & {60.00} & {65.00} & {123.33} & {433.33} \\
             \textbf{1.0} & \textbf{195} & \textbf{75.00} & \textbf{73.33} & \textbf{135.00} & \textbf{478.33} \\
             1.5 & {195} & {75.00} & {73.33} & {135.00} & {478.33} \\
            \bottomrule
        \end{tabular}
        }
        \end{small}
        \end{center}
    \end{minipage}
    \hfill
    \begin{minipage}[t!]{0.32\textwidth}
        \begin{center}
        \begin{small}
        \setlength\tabcolsep{3pt} 
        \scalebox{0.74}{
        \begin{tabular}{lx{25}x{25}x{25}x{25}c}
            \toprule
             \multicolumn{6}{c}{\textbf{\normalsize (c) LLaVA-1.5~\citep{liu2023improved} ($\lambda=1$)}} \\
             \arrayrulecolor{gray} \midrule
             & \multicolumn{2}{c}{\textbf{Object}} & \multicolumn{2}{c}{\textbf{Attribute}} & \multicolumn{1}{c}{\multirow{2}{*}{\textbf{\makecell{Total \\ Score}}}} \\
             \arrayrulecolor{gray} \cmidrule(lr){2-3} \cmidrule(lr){4-5} 
             $\alpha$ & Exist. & Count & Position & Color & \\
             \arrayrulecolor{gray} \midrule
             0.5 & {185} & {111.66} & {103.33} & {115.00} & {514.99} \\
             2.0 & {180} & {103.33} & {101.66} & {120.00} & {504.99} \\
             \textbf{2.5} & \textbf{180} & \textbf{105.00} & \textbf{111.66} & \textbf{120.00} & \textbf{516.66} \\
             3.0 & {180} & {105.00} & {111.66} & {120.00} & {516.66} \\
            \bottomrule
        \end{tabular}
        }
        \end{small}
        \end{center}
    \end{minipage}
    \vspace{\abovetabcapmargin}
    \caption{
        \textbf{$\alpha$ and $\lambda$ ablations on MME-Hallucination~\citep{fu2024mme}.}
        We set $\alpha=3$, $\lambda=1$ for InstructBLIP~\citep{dai2024instructblip} and $\alpha=2.5$, $\lambda=1$ for LLaVA-1.5~\citep{liu2023improved}. 
    }
    \label{tab:ablation}
\end{table*}

\vspace{\subsecmargin}\subsection{Ablation Study}
\paragraph{Ablations on $\alpha$ and $\lambda$.}
In our approach, $\lambda$ is the threshold for identifying blind tokens that receive excessive attention (see~\cref{eq:blind_token_identification}), and $\alpha$ controls the strength of contrastive decoding (see~\cref{eq:contrastive_decoding}). 
We conducted ablation experiments on the MME-Hallucination benchmark~\citep{liu2023mmbench} to study their effects.
\cref{tab:ablation} (a) and (c) show results using InstructBLIP~\citep{dai2024instructblip} and LLaVA-1.5~\citep{liu2023visual}, respectively, with $\lambda$ fixed at 1 and $\alpha$ varied from 0.5 to 3.
Overall, performance consistently improves with higher $\alpha$ values, with InstructBLIP achieving the highest total score at $\alpha$=3 and LLaVA-1.5 at $\alpha$=2.5.
These findings suggest that a stronger contrastive signal can better mitigate hallucinations. Additionally, \cref{tab:ablation} (b) shows that performance for InstructBLIP improves as $\lambda$ increases, indicating that restricting the application of our method to a smaller set of highly attended tokens yields better results.

\paragraph{Ablations on $\gamma$.}
We further evaluated the sensitivity of our approach to the parameter $\gamma$, which determines the cumulative threshold for selecting layers based on image attention (see \cref{eq:layer_selection}). Using LLaVA-1.5 with $\lambda$=1.0 and $\alpha$=2.5, our experiments (shown in \cref{table:gamma_ablation}) reveal that performance remains robust across a range of $\gamma$ values, except for extreme settings (e.g., $\gamma$=0.1). Our default value of $\gamma$=0.5 yields high accuracy and balanced metrics on the POPE-COCO-Random benchmark, and achieves the highest total score on the MME-Hallucination benchmark. Overall, these results indicate that our approach is not highly sensitive to $\gamma$, thereby reducing the need for extensive parameter tuning. Consequently, we fixed $\lambda$=1.0 and $\gamma$=0.5 in our experiments.


\paragraph{Layer Selection.}
We performed an ablation study to evaluate the effectiveness of our targeted layer selection strategy, which selects layers with the highest proportion of image-related attention for blind-token localization and calibration. To assess the contribution of this mechanism, we compared our method with three variants that manually fixed the selected layers (early, middle, or late five layers in the network).
As shown in \cref{table:layer_selection_ablation}, our approach consistently outperforms all alternatives across both the POPE-COCO-Random and MME-Hallucination benchmarks. These results demonstrate that our targeted layer selection yields tangible improvements, albeit with modest margins, indicating that selecting layers with stronger image-related attention leads to more accurate blind-token localization and reduces hallucination.

\begin{table}[t!]
    
    \centering
    \resizebox{0.7\linewidth}{!}{
        \begin{tabular}{ccccc}
        \hline
        \multicolumn{5}{c}{\textbf{(a) POPE-COCO-Random}} \\ \hline
        \textbf{$\gamma$} & \textbf{Acc.\up} & \textbf{Prec.\up} & \textbf{Rec.\up} & \textbf{F1\up} \\ \hline
        \textbf{0.5 (Ours)} & \colorbox{SecondBest}{87.93} & \colorbox{Best}{88.24} & 87.53 & \colorbox{SecondBest}{87.88} \\ 
        0.1 & 86.77 & 83.98 & \colorbox{Best}{90.87} & 87.29 \\ 
        0.3 & 87.47 & 85.35 & \colorbox{SecondBest}{90.47} & 87.83 \\ 
        1.0 & \colorbox{Best}{88.27} & \colorbox{SecondBest}{88.06} & 88.53 & \colorbox{Best}{88.30} \\
        \hline
        \end{tabular}
    }
    \vfill
    \vspace{2mm}
    \resizebox{\linewidth}{!}{
        \begin{tabular}{cccccc}
        \hline
        \multicolumn{6}{c}{\textbf{(b) MME-Hallucination}} \\ \hline
        \textbf{$\gamma$} & \textbf{Existence\up} & \textbf{Count\up} & \textbf{Position\up} & \textbf{Color\up} & \textbf{Total Score\up} \\ \hline
        \textbf{0.5 (Ours)} & \colorbox{Best}{189.29} & \colorbox{SecondBest}{104.76} & 106.19 & \colorbox{Best}{127.86} & \colorbox{Best}{528.10} \\ 
        0.1 & 167.50 & 101.80 & 103.33 & 117.50 & 490.13 \\ 
        0.3 & 180.00 & 98.33 & \colorbox{Best}{114.16} & \colorbox{SecondBest}{125.00} & 517.49 \\ 
        1.0 & \colorbox{SecondBest}{182.50} & \colorbox{Best}{108.33} & \colorbox{SecondBest}{109.99} & 117.50 & \colorbox{SecondBest}{518.32} \\ \hline
        \end{tabular}
    }
    \vspace{\abovetabcapmargin}
    \caption{
        \textbf{$\gamma$ ablations} on \textbf{(a)} POPE-COCO-Random and \textbf{(b)} MME-Hallucination benchmarks with LLaVA-1.5 ($\lambda$ = 1, $\alpha$ = 2.5).
    }
    \vspace{\belowtabcapmargin}
    \vspace{-3mm}
    \label{table:gamma_ablation}
\end{table}

\begin{table}[t!]
    \centering
    \resizebox{0.8\linewidth}{!}{
        \begin{tabular}{ccccc}
        \hline
        \multicolumn{5}{c}{\textbf{(a) POPE-COCO-Random}} \\ \hline
        \textbf{Method} & \textbf{Acc.\up} & \textbf{Prec.\up} & \textbf{Rec.\up} & \textbf{F1\up} \\ \hline
        \textbf{w/ layer selection (Ours)} & \colorbox{Best}{87.93} & \colorbox{SecondBest}{88.24} & \colorbox{Best}{87.53} & \colorbox{Best}{87.88} \\ 
        \arrayrulecolor{gray} \midrule
        Early 5 layers & 87.50 & 87.99 & 86.27 & 87.12 \\ 
        Mid 5 layers & \colorbox{SecondBest}{87.53} & 88.20 & 86.07 & 87.13 \\ 
        Last 5 layers & 87.46 & \colorbox{Best}{88.56} & 85.47 & 87.00 \\
        \hline
        \end{tabular}
    }
    \vfill
    \vspace{2mm}
    \resizebox{\linewidth}{!}{
        \begin{tabular}{cccccc}
        \hline
        \multicolumn{6}{c}{\textbf{(b) MME-Hallucination}} \\ \hline
        \textbf{Method} & \textbf{Existence\up} & \textbf{Count\up} & \textbf{Position\up} & \textbf{Color\up} & \textbf{Total Score\up} \\ \hline
        \textbf{w/ layer selection (Ours)} & \colorbox{Best}{189.29} & 104.76 & \colorbox{Best}{106.19} & \colorbox{Best}{127.86} & \colorbox{Best}{528.10} \\
        \arrayrulecolor{gray} \midrule
        Early 5 layers & 180.00 & \colorbox{Best}{108.33} & \colorbox{SecondBest}{106.66} & 120.00 & \colorbox{SecondBest}{514.99} \\ 
        Mid 5 layers & 180.00 & \colorbox{Best}{108.33} & 105.00 & 120.00 & 513.33 \\ 
        Last 5 layers & \colorbox{SecondBest}{185.00} & 103.33 & 105.00 & 120.00 & 513.33 \\ \hline
        \end{tabular}
    }
    \vspace{\abovetabcapmargin}
    \caption{
        \textbf{Ablation of layer selection} on \textbf{(a)} POPE-COCO-Random and \textbf{(b)} MME-Hallucination benchmarks. Our targeted layer selection outperforms manual alternatives in all key metrics.
    }
    \label{table:layer_selection_ablation}
\end{table}




\vspace{\secmargin}\section{Related Work}\vspace{\secmargin}
\label{sec:related_work}


To \textbf{mitigate hallucinations in LVLMs}, researchers have developed strategies across three levels:

\noindent\textbf{Input-level.}
These methods improve data quality and diversity by incorporating negative~\citep{liu2023mitigating} and counterfactual data~\citep{yu2023hallucidoctor} or through dataset cleansing~\citep{wang2024mitigating,yue2024less}, thereby fostering more robust visual-text alignments during training.

\noindent\textbf{Model-level.}
Approaches at this level enhance visual representations by increasing image processing resolution~\citep{chen2023internvl,liu2023improved,zhai2023halle} or by leveraging advanced vision encoders~\citep{he2024incorporating,jain2023vcoder,tong2024eyes}. Typically, these methods involve additional training with auxiliary supervision or reinforcement learning~\citep{zhao2023beyond,gunjal2024detecting,sun2023aligning,yu2023rlhf}.

\noindent\textbf{Output-level.}
Output-level methods directly refine the generated outputs. Contrastive decoding techniques~\citep{leng2023mitigating,favero2024multi} mitigate hallucinations by contrasting outputs from original and modified visual inputs, while guided decoding leverages external models like CLIP~\citep{radford2021learning} or DETR~\citep{carion2020end} to steer generation. Other approaches include training-free methods~\citep{wan2024contrastive,zhang2024debiasing,huang2023opera} and post-hoc corrections~\citep{lee2023volcano,wu2024logical}.
%
%

Our work falls within the output-level category. Unlike prior contrastive decoding methods that contrast whole-image representations, \Ours analyzes the internal attention patterns of LVLMs to identify \textit{blind tokens}---tokens that attract excessive attention but contribute little to the final output---and applies a contrastive decoding strategy to recalibrate their influence.

\vspace{\secmargin}\section{Conclusion}\vspace{\secmargin}
\label{sec:conclusion}
We identify and characterize blind tokens in LVLMs—image tokens that receive excessive attention while conveying little task-relevant information. These tokens misdirect the model’s focus, increasing the likelihood of hallucinated responses. To address this, we propose \OursFullAbb, a novel, training-free decoding technique that dynamically detects and mitigates the effect of blind tokens using image-wise attention analysis and contrastive decoding. Extensive evaluations on hallucination benchmarks demonstrate that \Ours improves both visual grounding and response accuracy, surpassing existing decoding strategies.


\vspace{\secmargin}\section*{Limitations}\vspace{\secmargin}
While \Ours reduces hallucinations, its effectiveness declines in tasks requiring precise object counting (e.g., the “Count” category in MME and “Number” in AMBER; see~\cref{tab:MME_full_apendix,tab:AMBER_discriminative})
This suggests that blind tokens may sometimes carry essential information for quantification.
\Ours adds some overhead due to dynamic test-time recalibration but maintains competitive tokens-per-second throughput compared to other contrastive decoding methods. Unlike high-latency beam search approaches (e.g., OPERA), \Ours offers a better efficiency–accuracy trade-off. Though currently sequential, recalibration can be parallelized to reduce wall-clock time at the expense of memory.
Performance on MME and AMBER varies with dataset scope and evaluation protocols, which are sensitive to token usage. Still, \Ours consistently lowers hallucination rates and yields statistically significant gains.



\paragraph{Future work.}
Building on insights from~\citep{darcet2023vision}, we hypothesize that blind token phenomenon may be intrinsic to large-scale transformer architectures, not limited to LVLMs. Future research will further explore these blind tokens and develop strategies to address them while balancing computational efficiency.

\bibliography{custom}
\clearpage

\addtocontents{toc}{\protect\setcounter{tocdepth}{2}}
\definecolor{linkcolor}{HTML}{000000}
\newpage
\section*{\centering\LARGE Appendix}
\label{sec:appendix}
\tableofcontents
\newpage
\appendix
\definecolor{linkcolor}{HTML}{ED1C24}

\vspace{10mm}
\section{Visualizations \& Analysis on Blind Tokens}
\label{sec:appendix_visualizations}
In this section, we provide a comprehensive analysis of the attention biases observed in LVLMs through extensive visualizations. Our findings reveal that LVLMs tend to allocate excessive attention to certain image tokens---termed \textit{blind tokens}---which, despite receiving high attention weights, contribute little to the final prediction logits.

\paragraph{More examples of attention bias.}
\label{sec:appendix_attention_bias}
\begin{figure*}[ht!]
    \centering
    \includegraphics[width=\linewidth]{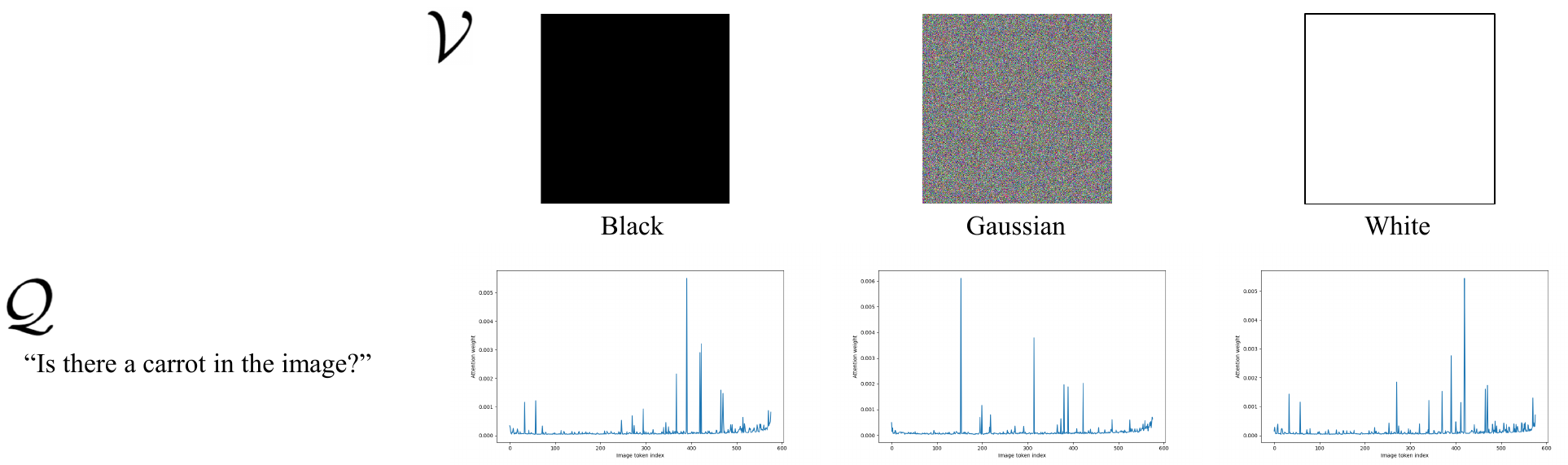} 
    \vspace{-6mm}
    \caption{
    \textbf{Attention distribution for images that lack semantic or query-relevant information.}
    Despite the absence of meaningful content, the model still focuses on certain regions, illustrating how blind tokens can dominate attention even in non-informative scenarios.
    }
    \vspace{-2mm}
    \label{fig:more_attention_bias}
\end{figure*}
To illustrate the phenomenon, we present several examples using LLaVA-1.5-7B. As shown in~\cref{fig:more_attention_bias}, the figure depicts three images---black, Gaussian noise, and white---lacking any semantic or query-related information. Despite the question “Is there a carrot in the image?” and the absence of any meaningful objects or features, the model still concentrates its attention on certain regions. This highlights how blind tokens can dominate the attention mechanism, even when there are no informative cues present. Such behavior demonstrates the tendency of LVLMs to latch onto seemingly random patches in the absence of salient visual details. We define blind tokens as image tokens that draw excessive attention while contributing little to the final prediction logits.

\vspace{1mm}\noindent\textbf{Visualization of blind tokens and target objects.}
\label{sec:appendix_visualization_blind_token}
\begin{figure}[ht!]
    \centering
    \vspace{-3mm}
    \includegraphics[width=\linewidth]{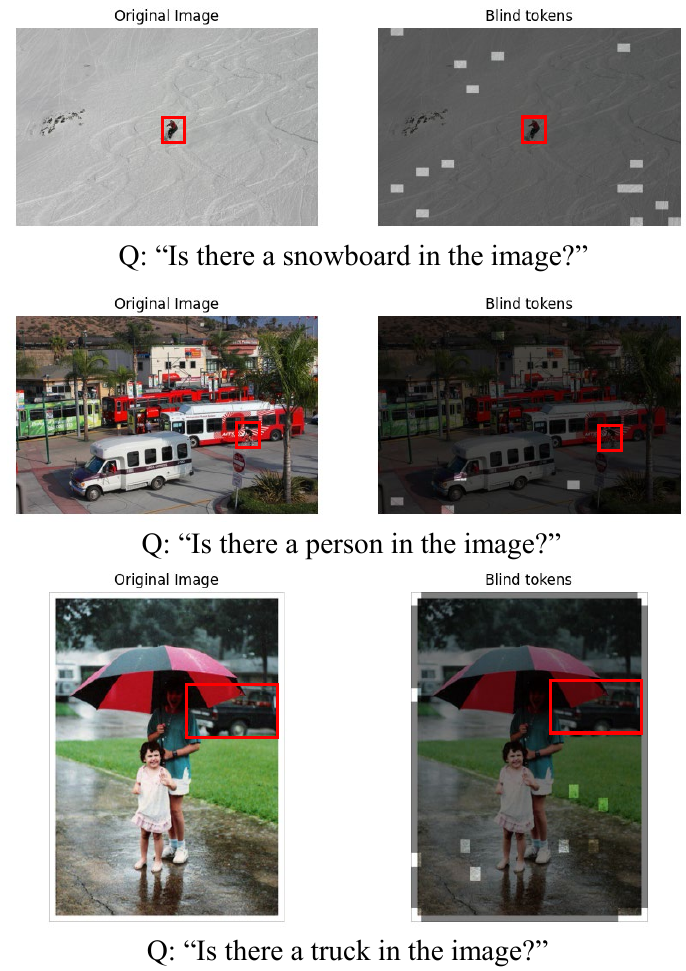} 
    \vspace{-6mm}
    \caption{
    \textbf{Visualization of blind tokens in real images from the POPE-COCO benchmark.} Each row displays \textbf{(left)} the original image and \textbf{(right)} the same image with blind tokens highlighted. The red boxes indicate areas where query-related objects are located. 
    }
    \label{fig:blind_token_pope}
\end{figure}
In \cref{fig:blind_token_pope}, we overlay the locations of blind tokens with bounding boxes of target objects on images from the POPE-COCO benchmark. This visualization supports our claim that there is a mismatch between the highly attended blind tokens and the regions containing query-relevant information.

\vspace{1mm}\noindent\textbf{Distributions of bounding boxes and blind tokens.}
\label{sec:appendix_heatmap}
\begin{figure}[ht!]
    \centering
    \vspace{-3mm}
    \includegraphics[width=\linewidth]{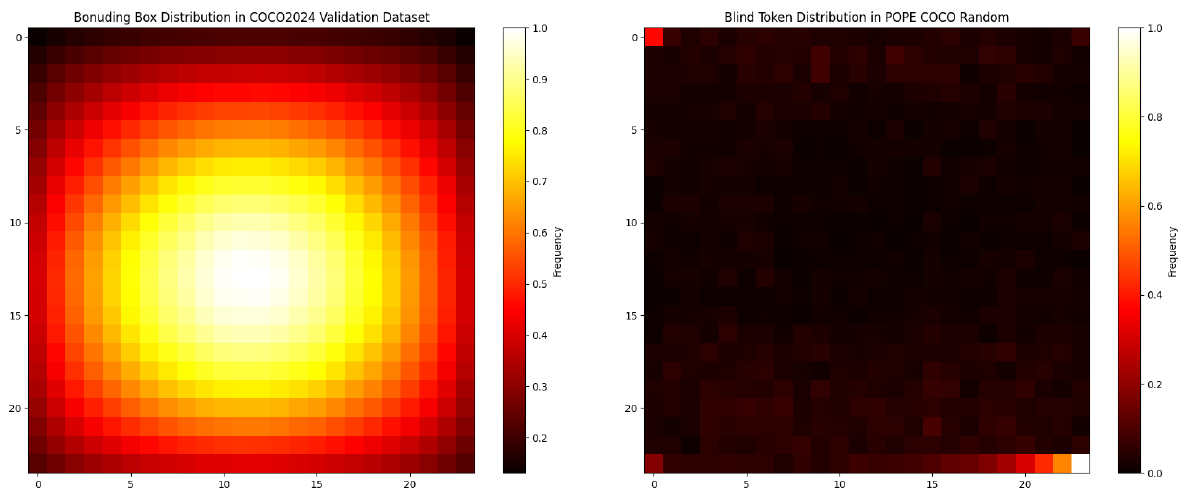} 
    \vspace{-6mm}
    \caption{
    \textbf{Comparison of (left) bounding box distribution with (right) the distribution of blind tokens in the COCO dataset.} Warmer colors indicate higher density, revealing that bounding boxes cluster near the center while blind tokens are more prevalent around the periphery. This highlights a spatial mismatch between regions containing genuine objects and areas receiving disproportionately high attention.
    }
    \vspace{-3mm}
    \label{fig:dist_heat_map}
\end{figure}
Heatmaps in \cref{fig:dist_heat_map} illustrate that while object bounding boxes tend to be centered, blind tokens are predominantly located along the image edges, revealing a significant spatial disparity.

\vspace{1mm}\noindent\textbf{Visualization and statistics of blind tokens.}
\label{sec:appendix_statistics_blind_token}
\begin{figure*}[ht!]
    \centering
    \vspace{-2mm}
    \includegraphics[width=\linewidth]{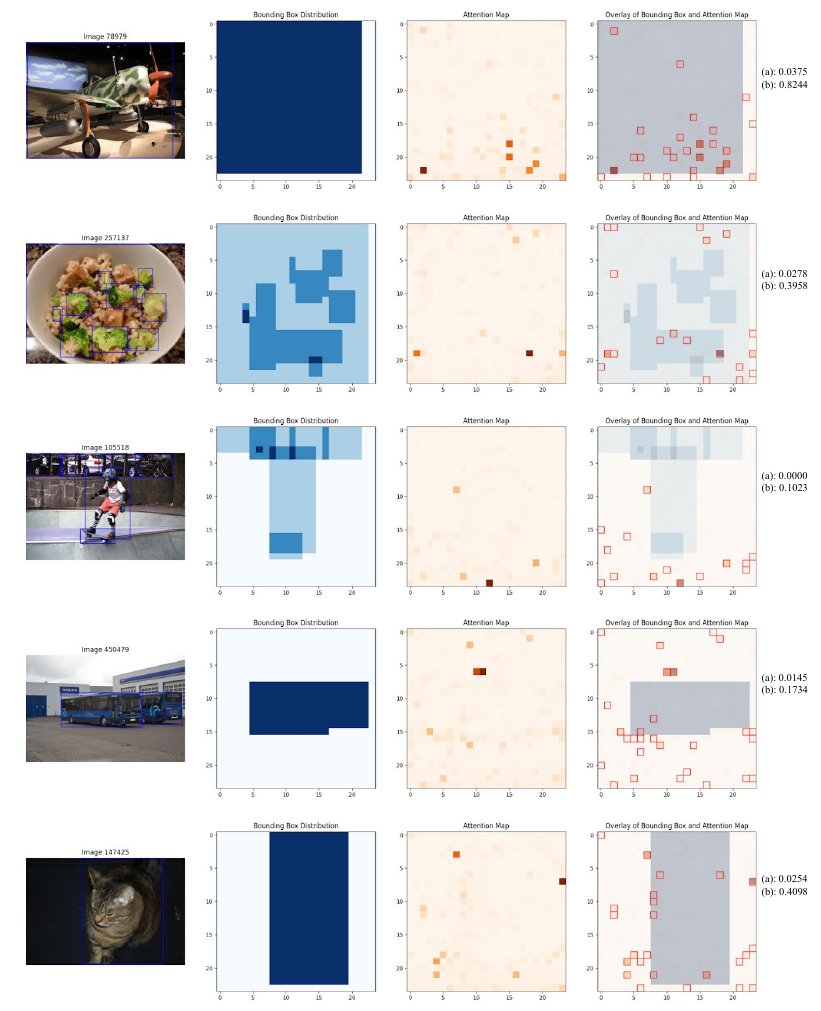} 
    \vspace{1em}
    \begin{tabular}{|c|c|}
        \hline
         (a) Avg. \# blind tokens in BBox / \# BBox tokens  &  3.68\%\\
         \hline
         (b) Avg. BBox token attention propotion & 23.2\% \\
         \hline
    \end{tabular}
    \vspace{-5mm}
    \caption{
    \textbf{Visualization and statistics of object bounding boxes and blind tokens in the COCO2014 dataset.}
    Each row displays (from left to right) the original image, bounding box distribution, attention map, and blind tokens (in red boxes). On average, only 3.68\% of blind tokens overlap with bounding boxes, while bounding box regions receive just 23.2\% of the total attention. This highlights a clear mismatch between regions containing genuine objects and those receiving high attention.
    }
    \vspace{-2mm}
    \label{fig:blind_token_statistics}
\end{figure*}
We conducted a correlation analysis on 3,000 images from the COCO2014 validation dataset. The results are in~\cref{fig:blind_token_statistics}. LVLMs were tasked with describing images, and we analyzed the attention distribution over 24 $\times$ 24 patches. Our results indicate that, on average, only 3.7\% of blind tokens overlap with actual object regions, with merely 23.3\% of the total attention weight allocated to these regions---highlighting the disconnect between blind tokens and task-relevant information.

\vspace{1mm}\noindent\textbf{Histogram of blind tokens.}
\label{sec:appendix_histogram_blind_token}
\begin{figure*}[ht!]
    \centering
    \includegraphics[width=\linewidth]{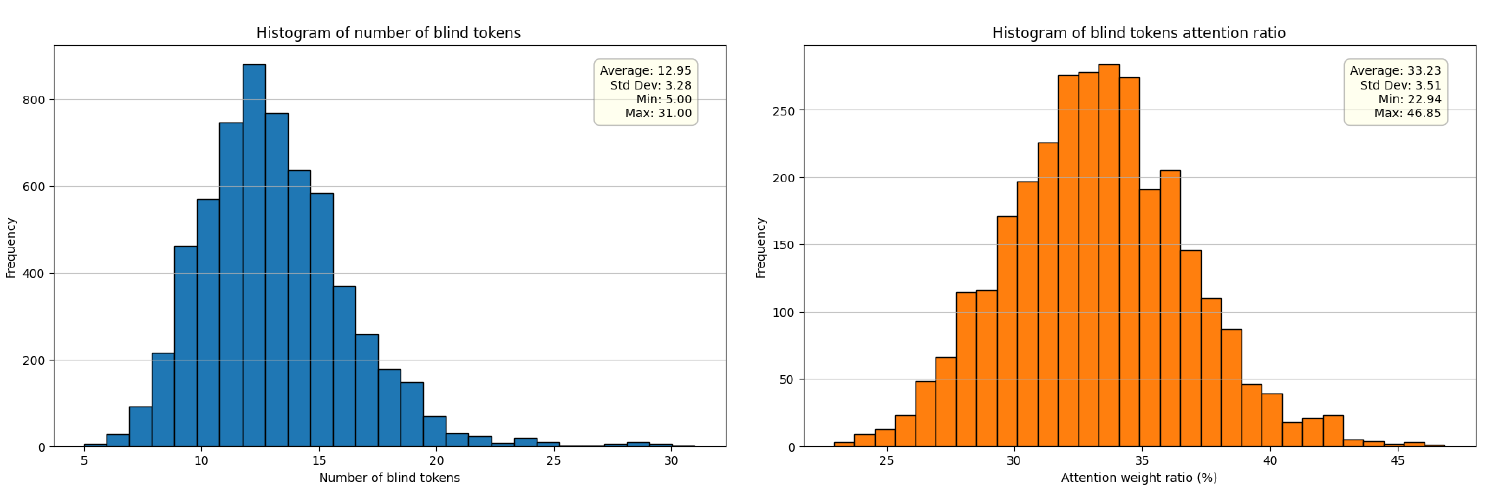} 
    \vspace{-7mm}
    \caption{
    \textbf{Histograms illustrating the distribution of blind tokens in the POPE-COCO-Random benchmark.}
    The left histogram shows the average number of blind tokens per image (about 13), while the right histogram indicates that these tokens account for roughly 33\% of the total attention weight.
    This highlights the disproportionate influence blind tokens exert on the model’s attention.
    }
    \vspace{-2mm}
    \label{fig:histogram}
\end{figure*}
\cref{fig:histogram} presents a histogram of the number of blind tokens and their corresponding attention weights. In our evaluation with LLaVA-1.5-7B on the POPE-COCO-Random benchmark, we identified an average of 12.95 blind tokens, which accounted for 33.23\% of the total image token attention weight.

\vspace{1mm}\noindent\textbf{Blind tokens and token probability distribution.}
\label{sec:appendix_prob_dist_blind_token}
\begin{figure*}[ht!]
    \centering
    \includegraphics[width=\linewidth]{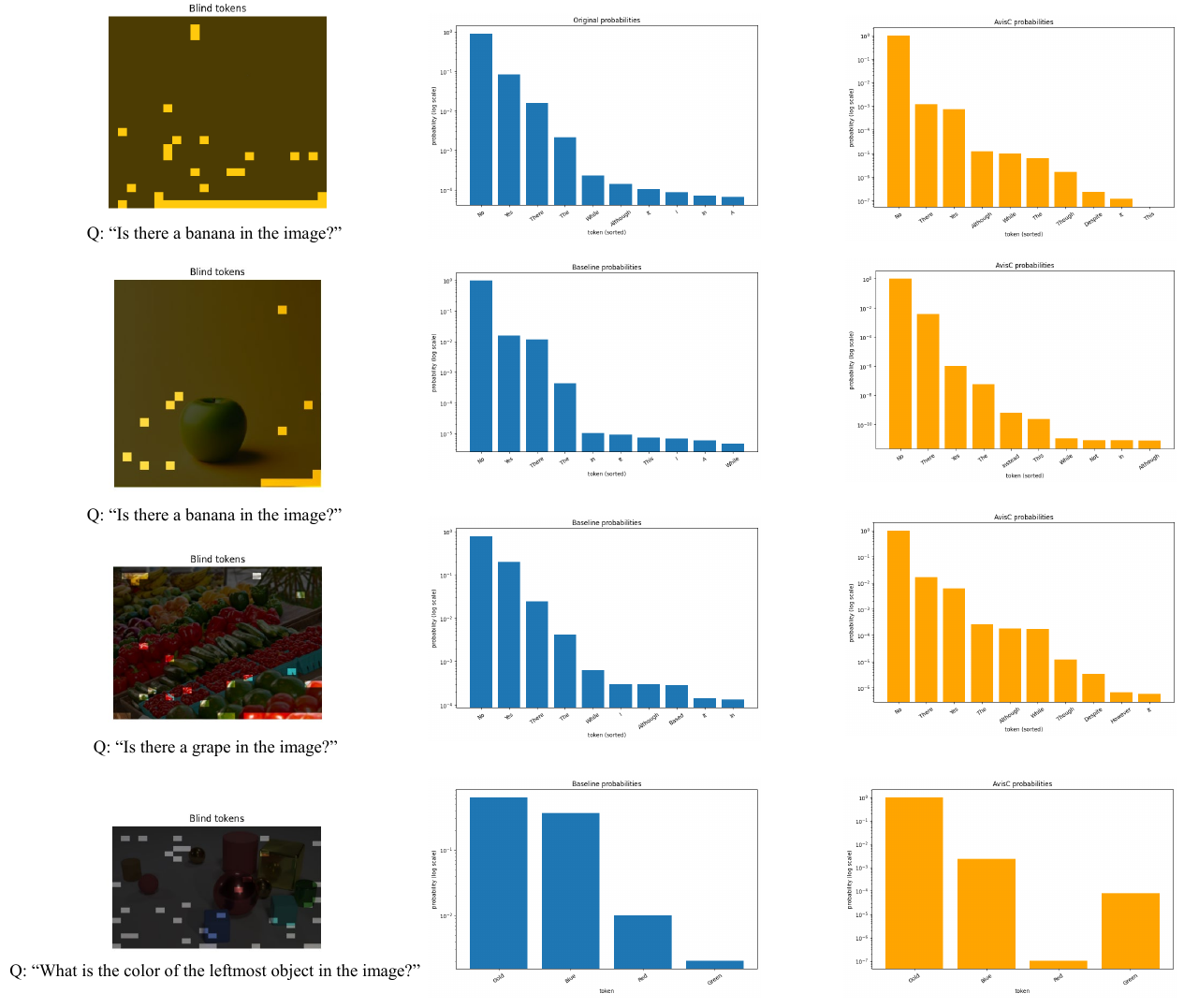} 
    \vspace{-9mm}
    \caption{
    \textbf{Visualization of blind tokens and the logit probability distributions before and after \Ours.} Each row shows \textbf{(left)} the image with blind tokens highlighted, \textbf{(center)} the model’s original prediction logits, and \textbf{(right)} the logits adjusted by \Ours. By recalibrating attention away from blind tokens, \Ours increases the accuracy and confidence of the model’s responses to queries.
    }
    \vspace{-2mm}
    \label{fig:blind_token_prob}
\end{figure*}
\cref{fig:blind_token_prob} visualizes the location of blind tokens for a given image and query, and presents the token logit values of both the baseline model and \Ours.
For example, in the first problem, which asked whether there is a banana in the image, the original probability distribution was: 'No' at 89.62\%, 'Yes' at 8.46\%, and 'There' at 1.56\%. After applying \Ours, the logit distribution shifted to: 'No' at 98.00\%, 'There' at 1.35\%, and 'Yes' at 0.61\%.

\section{More Experimental Details}
\label{sec:appendix_experimental_details}

\subsection{Further Implementation details}
\label{sec:appendix_implementation_details}

Our decoding process employs cut-off sampling following VCD~\citep{leng2023mitigating}.
Tokens with probability below $\beta$ times the maximum probability at each generation step are masked.
Formally, we consider text tokens $\xi_t \in \mathcal{H}$ satisfying:
\begin{equation}
    \mathcal{H}(\xi_{<t}) = \left\{ \xi_t \in \mathcal{H} \;\middle|\;
    p\left(\xi_t \mid \mathcal{V}, \mathcal{Q}, \xi_{<t} ; \theta\right) 
    \right\} 
\end{equation}
\begin{equation*}
    \quad \geq \beta \max _w p\left(w \mid \mathcal{V}, \mathcal{Q}, \xi_{<t} ; \theta\right).
    \label{eq:appendix_ritual_cd_sampling}
\end{equation*}
We set $\beta$=0.1 and limit generation to a maximum of 64 tokens per task.
For LLaVA-1.5~\citep{liu2023visual} experiments, we used the \texttt{llava}$_\texttt{v1}$ conversation template.

For reproducing VCD~\citep{leng2023mitigating}, we followed the official code with $\alpha$ = 1.0, $\beta$ = 0.1, and a diffusion noise step $T$ = 500. In our M3ID~\citep{favero2024multi} reproduction, we set $\lambda$ = 0.2. These settings ensure fair comparisons across methods.

\subsection{Evaluation Benchmarks}

\paragraph{POPE.}
We utilize the official POPE benchmark~\citep{li2023evaluating}, which includes 3,000 question-answer pairs (across random, popular, and adversarial setups) with queries of the form “Is there a [object] in the image?”. Performance is measured by accuracy, precision, recall, and mean F1-score.\footnote{\href{https://github.com/RUCAIBox/POPE}{https://github.com/RUCAIBox/POPE}}

\paragraph{MME.}
The MME dataset~\citep{fu2024mme} is divided into 10 perceptual categories (existence, count, position, color, posters, celebrity, scene, landmark, artwork, OCR) and 4 cognitive categories (commonsense reasoning, numerical calculation, text translation, code reasoning). 
We use the official dataset but remove the one-word response constraint to allow natural responses.\footnote{\href{https://github.com/BradyFU/Awesome-Multimodal-Large-Language-Models/tree/Evaluation}{https://github.com/BradyFU/Awesome-Multimodal-Large-Language-Models/tree/Evaluation}} 

\paragraph{AMBER.}
AMBER~\citep{wang2023llm} comprises 1004 images with both generative (e.g., “Describe this image.”) and discriminative (existence, attribute, relation) tasks. We randomly sample 500 questions for generative and 5000 for discriminative tasks, following official protocols.
\footnote{\href{https://github.com/junyangwang0410/AMBER.git}{https://github.com/junyangwang0410/AMBER.git}}

\paragraph{LLaVA-Bench.}
LLaVA-Bench~\citep{liu2023visual} features 24 images and 60 questions covering diverse contexts (e.g., indoor, outdoor, paintings, sketches) to test LVLM adaptability.\footnote{\href{https://huggingface.co/datasets/liuhaotian/llava-bench-in-the-wild}{https://huggingface.co/datasets/liuhaotian/llava-bench-in-the-wild}}

\subsection{Metrics}
\paragraph{Metrics on the MME.}
For each visual input $\mathcal{V}$ and its discriminative questions \{$q_1, q_2$\}, we define accuracy (\textit{ACC}) as:
\begin{equation}
    \text{\textit{ACC}}(\mathcal{V}, q_i) = \begin{cases}
  1 & \text{if LVLMs$(\mathcal{V}, q_i)$} \\
    & \quad = \text{Answer$(\mathcal{V}, q_i)$}, \\
  0 & \text{otherwise} .
\end{cases}
\end{equation}
An additional metric, \textit{ACC+}~\citep{fu2024mme}, is 1 if both answers for an image are correct, and 0 otherwise.
\begin{equation}
    \text{\textit{ACC+}}(\mathcal{V}) = \begin{cases}
  1 & \text{if LVLMs$(\mathcal{V}, q_i)$} \\
    & \quad = \text{Answer$(\mathcal{V}, q_i)$ for any $i$}, \\
  0 & \text{otherwise} .
\end{cases}
\end{equation}
The overall MME score is the sum of \textit{ACC} and \textit{ACC+}.

\paragraph{Metrics on the generative tasks.}
Let R denote the response generated for a visual input V. We employ:

\noindent\textbf{(1) \textit{CHAIR}}~\citep{rohrbach2018object, wang2023llm} evaluates the occurrence of hallucinatory objects in responses to LVLMs. 
\textit{CHAIR} uses an annotated list of objects $A$=\{$a_{obj}^1$, $a_{obj}^2$, $\ldots$, $a_{obj}^n$\} to calculate how often hallucinated objects appear in the responses. 
Let $R$=\{$r_{obj}^1$, $r_{obj}^2$, $\ldots$, $r_{obj}^m$\} be the list of objects mentioned in the response of LVLMs, the formula for \textit{CHAIR} is given as:
\begin{equation}
    \text{\textit{CHAIR}} = 1 - \frac{len(R\cap A)}{len(R)}.
\end{equation}

\noindent\textbf{(2) \textit{Cover}}~\citep{wang2023llm} 
The \textit{Cover} metric measures how completely the objects in the response cover the identified objects in the image. 
\textit{Cover} calculates the ratio of objects mentioned in the response to the total objects listed.
The formula for \textit{Cover} is:
\begin{equation}
    \text{\textit{Cover}} = \frac{len(R \cap A)}{len(A)}.
\end{equation}

\noindent\textbf{(3) \textit{Hal}}~\citep{wang2023llm}
The \textit{Hal} metric quantifies the presence of hallucinations by checking if the \textit{CHAIR} value is not zero, indicating the presence of hallucinations. 
The \textit{Hal} is presented by the following formula:
\begin{equation}
    \text{\textit{Hal}} = \begin{cases}
  1 & \text{if $\text{\textit{CHAIR}} \neq 0$}, \\
  0 & \text{otherwise} .
\end{cases}
\end{equation}

\noindent\textbf{(4) \textit{Cog}}~\citep{wang2023llm}
The \textit{Cog} metric evaluates whether the hallucinations in LVLMs responses resemble human cognition.
The \textit{Cog} calculates the ratio of the human hallucinatory object targets, denoted as $H$=\{$h_{obj}^1$, $h_{obj}^2$, $\ldots$, $h_{obj}^n$\} to the objects mentioned in the response.
The formula for \textit{Cog} is: 
\begin{equation}
    \text{\textit{Cog}} = \frac{len(R \cap H)}{len(R)}.
\end{equation}

\noindent\textbf{(5) \textit{AMBER Score}~\citep{wang2023llm}}
The \textit{AMBER Score} metric evaluates the comprehensive performance of LVLMs for generative tasks and discriminative tasks. 
This score combines the CHAIR metric for generative tasks with the F1 metric for discriminative tasks.
The formula representing the \textit{AMBER Score} is as follows:
\begin{equation}
    \text{\textit{AMBER Score}} = \frac{1}{2}\times(1 - \textit{CHAIR} + \textit{F1}).
\end{equation}

\section{Additional Experiments}
\label{sec:apependix_additional_experiments}

\begin{table*}[t!]
    \centering
    \small
    \setlength\tabcolsep{5pt}
    \scalebox{1}{
        \begin{tabular}{x{60}x{30}x{40}x{75}x{25}x{25}x{61}}
        \toprule
        \textbf{Case}        & \textbf{\# Case (Base)} & \textbf{\# Case (\Ours)} & \textbf{Logit Yype}           & \textbf{"Yes" Logit} & \textbf{"No" Logit} & \textbf{GT Logit - Wrong Logit{\up}} \\ \hline
        \multirow{3}{*}{TP (GT = Yes){\up}} & \multirow{3}{*}{3952} & \multirow{3}{*}{3958 {\color{plus}(+6)}} & Baseline              & 30.34            & 25.68            & 4.67                      \\ 
                                  &                        &                       & Zero-out > $\mu$ + $\sigma$       & 28.78            & 25.56            & 3.22                      \\  
                                  &                        &                       & Zero-out < $\mu$ + $\sigma$   & 20.14            & 19.40            & 0.74                      \\ \hline
        \multirow{3}{*}{TN (GT = No){\up}}  & \multirow{3}{*}{3317} & \multirow{3}{*}{3536 {\color{plus}(+219)}} & Baseline              & 26.60            & 28.88            & 2.28                      \\ 
                                  &                        &                       & Zero-out > $\mu$ + $\sigma$       & 25.82            & 28.45            & 2.63                      \\  
                                  &                        &                       & Zero-out < $\mu$ + $\sigma$   & 18.78            & 19.21            & 0.43                      \\ \hline
        \multirow{3}{*}{FP (GT = No) {\down}}  & \multirow{3}{*}{1183} & \multirow{3}{*}{964 {\color{minus}(-219)}}  & Baseline              & 28.01            & 27.61            & -0.40                     \\ 
                                  &                        &                       & Zero-out > $\mu$ + $\sigma$       & 26.75            & 27.48            & 0.74                     \\ 
                                  &                        &                       & Zero-out < $\mu$ + $\sigma$   & 19.33            & 19.29            & -0.04                     \\ \hline
        \multirow{3}{*}{FN (GT = Yes) {\down}} & \multirow{3}{*}{548}  & \multirow{3}{*}{542 {\color{minus}(-6)}}   & Baseline              & 27.42            & 28.05            & -0.63                     \\ 
                                  &                        &                       & Zero-out > $\mu$ + $\sigma$       & 26.41            & 27.76            & -1.36                     \\ 
                                  &                        &                       & Zero-out < $\mu$ + $\sigma$   & 19.08            & 19.26            & -0.18                     \\ \bottomrule
        \end{tabular}
    }
    \vspace{\abovetabcapmargin}
    \caption{
        \textbf{Zero-out experiments on the POPE-COCO benchmark~\cite{rohrbach2018object}.}
        We compare how logits change under two strategies: (1) zeroing out blind tokens (Zero-out $> \mu + \sigma$) and (2) zeroing out non-blind tokens (Zero-out $< \mu + \sigma$).
        Rows denote true positives (TP), true negatives (TN), false positives (FP), and false negatives (FN). For each case, we show the average “Yes” logit, “No” logit, and the difference between the ground‐truth (GT) logit and the wrong logit.
        All results are obtained with LLaVA‐1.5‐7B~\cite{liu2023improved} on 3,000 MS‐COCO~\citep{lin2014microsoft} images.
    }
    \label{tab:zero_out_experiments}
    \vspace{\belowtabcapmargin}
    \vspace{-2mm}
\end{table*}

\subsection{Zero-Out Experiments on POPE-COCO Benchmark}
In~\cref{tab:zero_out_experiments}, we compare how logits change under two strategies on the POPE-COCO benchmark~\cite{rohrbach2018object}: (1) zeroing out blind tokens (\ie, tokens with attention $> \mu + \sigma$) and (2) zeroing out non-blind tokens (\ie, tokens with attention $< \mu + \sigma$).
Here, removing blind tokens minimally alters the model’s predictions, indicating that they hold little object‐discriminative information. In contrast, removing non‐blind tokens drastically shifts the logits, underscoring their critical importance.
This indicates that blind tokens have a smaller impact on prediction logits than non-blind tokens.
Compared to base decoding, \Ours effectively reduces over-emphasis on blind tokens, improving performance, particularly for TN and FP cases.

\subsection{Inference Time and OPERA}
\label{sec:appendix_inference_time}
\begin{table}[t!]
    
    \centering
    \resizebox{\linewidth}{!}{
    \begin{tabular}{lccccc}
        \toprule
         Method & Acc.\up & Prec.\up & Rec.\up & F1\up & tokens/sec\up \\ \hline
        
         \textit{base} & 84.47 & 83.35 & 86.13 & 84.72 & \colorbox{Best}{24.44} \\  
         VCD & 84.80 & 83.00 & 87.53 & 85.20 & 11.53 \\  
         M3ID & 86.00 & 85.11 & 87.27 & 86.18 & \colorbox{SecondBest}{13.14} \\  
         \textbf{AvisC} & \colorbox{SecondBest}{87.93} & \colorbox{SecondBest}{88.24} & \colorbox{SecondBest}{87.53} & \colorbox{SecondBest}{87.88} & {12.28} \\ \midrule
         OPERA (Beam=2) & \colorbox{Best}{89.35} & \colorbox{Best}{90.37} & \colorbox{Best}{88.80} & \colorbox{Best}{89.58} & 0.17 \\  \hline
    \end{tabular}
    }
    \vspace{-2mm}
    \caption{
    \textbf{Comparison of inference time and performance on the POPE-COCO-Random benchmark for LLaVA-1.5.}
    While OPERA achieves the highest performance metrics, it operates at a substantially slower speed compared to the other methods.
    }
    \vspace{-3mm}
    \label{tab:inference_time}
\end{table}

\cref{tab:inference_time} presents an efficiency and performance comparison between contrastive decoding methods (\Ours, M3ID, OPERA, and VCD) and \Ours. Inference speed is measured with a TiTAN RTX GPU on the POPE-COCO-Random benchmark.
OPERA introduces the concept of an "anchor token" and uses this token to guide sentence generation and rollback, thereby mitigating hallucinations.
OPERA is implemented on the beam search decoding method of LLMs, so a fair comparison with \Ours is not possible. However, OPERA showed the best performance overall. However, its inference speed was approximately $\times$72.23 slower than \Ours.

\begin{table}[t!]
    \begin{center}
        \begin{small}
        \vspace{-2mm}
        \renewcommand{\arraystretch}{1.3} 
        \setlength\tabcolsep{4pt} 
        \scalebox{0.85}{
        \begin{tabular}{llx{25}x{25}x{25}x{25}}
            \toprule
             & \textbf{Case} & Acc.\up & Prec.\up & Rec.\up & F1\up\\
            \arrayrulecolor{gray} \midrule
            \multirow{4}{*}{\rot{\textbf{InstructBLIP}}} & Zeros & \colorbox{SecondBest}{88.50} & \colorbox{Best}{93.00} & 83.27 & \colorbox{SecondBest}{87.86} \\
             & Ones & {82.50} & {75.48} & \colorbox{Best}{96.27} & 84.62 \\
             & Noise & {86.77} & \colorbox{SecondBest}{84.71} & \colorbox{SecondBest}{89.73} & 87.15 \\
             & Mask & \colorbox{Best}{88.53} & 90.14 & 86.53 & \colorbox{Best}{88.30} \\
            \arrayrulecolor{gray} \midrule
            \multirow{4}{*}{\rot{\textbf{LLaVA 1.5}}} & Zeros & \colorbox{SecondBest}{87.87} & \colorbox{SecondBest}{88.12} & \colorbox{SecondBest}{87.53} & \colorbox{Best}{87.83} \\
             & Ones & 79.97 & 72.22 & \colorbox{Best}{97.40} & 82.94 \\
             & Noise & \colorbox{Best}{88.47} & \colorbox{Best}{93.19} & 83.00 & \colorbox{SecondBest}{87.80} \\
             & Mask & 84.77 & 86.29 & 82.67 & 84.44 \\
            \bottomrule
        \end{tabular}
        }
        \vspace{\abovetabcapmargin}
        \caption{
        \textbf{Design choices for non-blind image token deactivation.}
        Each row presents a different method for handling non-blind tokens (Zeros, Ones, Noise, or Mask), and shows the resulting performance.
        }
        \vspace{\belowtabcapmargin}
        \vspace{-5mm}
        \label{tab:zeroout_ablations}
        \end{small}
    \end{center}
\end{table}

\subsection{Alternatives to Zero-Out}
\Cref{tab:zeroout_ablations} shows ablation results for various deactivation schemes applied to non-blind image tokens on the POPE-COCO-random benchmark~\citep{li2023evaluating}, using both InstructBLIP~\citep{dai2024instructblip} and LLaVA 1.5~\citep{liu2023visual}. We compare four methods: setting tokens to zero (Zeros), to ones (Ones), replacing tokens with noise (Noise), and masking tokens out in the attention mechanism (Mask). For InstructBLIP, the Mask approach achieves the highest Accuracy and F1 score, while the Zeros method excels in Precision; Ones yields the best Recall, and Noise offers balanced performance across Precision and Recall. For LLaVA 1.5, Noise achieves the highest Accuracy and Precision, whereas Zeros demonstrates consistent, balanced performance across all metrics. Overall, the Zeros approach proved most effective in calibrating attention to image tokens and improving model performance.

\begin{table}[t!]
\centering
\resizebox{.9\linewidth}{!}{
\begin{tabular}{llcccc}
\toprule
\multicolumn{1}{l}{\multirow{2}{*}{Setup}} & \multicolumn{1}{l}{\multirow{2}{*}{Method}} & \multicolumn{4}{c}{\textbf{LLaVA-1.5 (13B)}}                                                      \\ \cmidrule(lr){3-6}
\multicolumn{1}{l}{}                       & \multicolumn{1}{l}{}                        & Acc.\up           & Prec.\up          & Rec.\up           & F1\up              \\ \hline
\multirow{5}{*}{Random}                    & \textit{base}                                    & 83.17          & \colorbox{SecondBest}{79.49}          & 89.40          & 84.15                    \\
                                           & VCD                                         & 82.97          & 78.90          & 90.00          & 84.09             \\
                                           & M3ID                                        & \colorbox{SecondBest}{83.43}    & 79.31    & \colorbox{SecondBest}{90.47}    & \colorbox{SecondBest}{84.52}          \\
                                           & \Ours                                      & \colorbox{Best}{88.40} & \colorbox{Best}{86.05} & \colorbox{Best}{91.67} & \colorbox{Best}{88.77}  \\ \hline
\multirow{5}{*}{Popular}                   & \textit{base}                                    & \colorbox{SecondBest}{80.93}     & \colorbox{SecondBest}{76.45}   &  89.40   & 82.42                    \\
                                           & VCD                                         & 79.67 &    74.59  &   90.00   & 81.57            \\
                                           & M3ID                                        & 80.90    & 75.94 &    \colorbox{SecondBest}{90.47} &   \colorbox{SecondBest}{82.57}   \\
                                           & \Ours                                      &\colorbox{Best}{85.73} &    \colorbox{Best}{81.94}  &   \colorbox{Best}{91.67}   & \colorbox{Best}{86.53}  \\ \hline
\multirow{5}{*}{Adversarial}               & \textit{base}                                    & \colorbox{SecondBest}{76.03}   &  \colorbox{SecondBest}{70.74}  &   88.80 &   78.75      \\
                                           & VCD                                         & 75.57     &69.86   &  89.93  &  78.64           \\
                                           & M3ID                                        & 75.80     &69.97     &\colorbox{SecondBest}{90.40}  &  \colorbox{SecondBest}{78.88} \\
                                           & \Ours                                      & \colorbox{Best}{79.27}&     \colorbox{Best}{73.65} &    \colorbox{Best}{91.13} &   \colorbox{Best}{81.47} 
\\
\bottomrule
\end{tabular}

}
\vspace{-2mm}
\caption{
\textbf{Results of LLaVA-1.5-13B on POPE-COCO benchmark.}
}
\vspace{-2mm}
\label{tab:13B models}
\end{table}

\begin{table}[t!]
\centering
\resizebox{\linewidth}{!}{
\begin{tabular}{llcccc}
\toprule
\multirow{2}{*}{Setup} & \multirow{2}{*}{Method} 
& \multicolumn{2}{c}{
    \begin{tabular}{@{}c@{}}\textbf{LLaVA-OneVision}\\(\textbf{Qwen2-7B})\end{tabular}
} 
& \multicolumn{2}{c}{\textbf{Qwen2.5-VL-7B}} \\
\cmidrule(lr){3-4} \cmidrule(lr){5-6}
& & Acc. & F1 & Acc. & F1 \\
\midrule
\multirow{4}{*}{Random}
  & \textit{base}  & 88.60 & 87.34 & 88.84 & 87.82 \\
  & VCD            & \colorbox{SecondBest}{90.57} & \colorbox{SecondBest}{89.71} & \colorbox{SecondBest}{90.47} & \colorbox{SecondBest}{89.37} \\
  & M3ID           & 89.87 & 88.91 & 90.77 & 88.57 \\
  & \Ours          & \colorbox{Best}{91.46} & \colorbox{Best}{90.84} & \colorbox{Best}{92.36} & \colorbox{Best}{91.50} \\
\midrule
\multirow{4}{*}{Popular}
  & \textit{base}  & 84.60 & 83.79 & 85.05 & 83.72 \\
  & VCD            & 87.20 & \colorbox{SecondBest}{86.50} & 87.10 & \colorbox{SecondBest}{86.21} \\
  & M3ID           & \colorbox{SecondBest}{87.60} & 85.65 & \colorbox{SecondBest}{88.50} & 85.36 \\
  & \Ours          & \colorbox{Best}{89.86} & \colorbox{Best}{89.45} & \colorbox{Best}{90.76} & \colorbox{Best}{90.16} \\
\midrule
\multirow{4}{*}{Adversarial}
  & \textit{base}  & 83.00 & 82.40 & 84.07 & 82.73 \\
  & VCD            & 86.00 & \colorbox{SecondBest}{85.42} & 86.50 & \colorbox{SecondBest}{85.52} \\
  & M3ID           & \colorbox{SecondBest}{86.80} & 84.86 & \colorbox{SecondBest}{87.30} & 84.96 \\
  & \Ours          & \colorbox{Best}{86.12} & \colorbox{Best}{85.83} & \colorbox{Best}{87.62} & \colorbox{Best}{86.93} \\
\bottomrule
\end{tabular}
}
\vspace{-2mm}
\caption{
\textbf{POPE (MS-COCO) results on LLaVA-OneVision-Qwen2-7B and Qwen2.5-VL-7B.}
}
\label{tab:recent_lvlms}
\vspace{-2mm}
\end{table}

\subsection{Results of Larger LVLM}
\cref{tab:13B models} presents the performance of each method on the POPE benchmark using the COCO dataset based on the LLaVA-1.5v-13B model. In this experiment setup, compared to the 7B small model shown in \cref{tab:POPE}, the performance improvement of \Ours is even more pronounced. For other methods (\ie, VCD, M3ID), the performance increase is slight or, in some cases, decreases depending on the metric. However, \Ours demonstrates robust performance improvement, remaining resilient to changes in the size of LVLMs.



\subsection{Additional Evaluation on off-ths-shelf LVLMs}
To further validate the model-agnostic robustness and generalizability of \Ours, we extended our evaluation to include recent and diverse LVLMs with varying attention mechanisms. Specifically, we tested \Ours on LLaVA-OneVision-Qwen2-7B, Qwen2.5-VL-7B. As shown in \cref{tab:recent_lvlms}, \Ours consistently outperforms prior methods across all setups on the POPE benchmark, achieving significant reductions in hallucination without compromising the original model capabilities. These results support \Ours’s effectiveness as a test-time, plug-and-play strategy applicable across off-the-shelf LVLMs.

\subsection{Additional Evaluation on Generative Benchmarks}
To more comprehensively evaluate the impact of \Ours on generative capabilities, we conducted additional experiments on two free-form generation benchmarks: MMHal-Bench and Object-Hallucination. The results, summarized in \cref{tab:generative_benchmark}, show that \Ours not only preserves the generative quality of the base models (e.g., maintaining or improving MMHal scores), but also consistently reduces hallucination metrics compared to existing approaches. These findings highlight that \Ours serves as a reliable and effective test-time method that retains the strengths of pretrained LVLMs in generative settings.

\begin{table}[t!]
\centering
\resizebox{\linewidth}{!}{
\begin{tabular}{llcccc}
\toprule
\multirow{2}{*}{\textbf{Model}} & \multirow{2}{*}{\textbf{Method}} 
& \multicolumn{2}{c|}{\textbf{MMHal-Bench}} 
& \multicolumn{2}{c}{\textbf{Object-Hallucination}} \\
\cmidrule(lr){3-4} \cmidrule(lr){5-6}
& & Score $\uparrow$ & HalRate $\downarrow$ & CHAIR$_s$ $\downarrow$ & CHAIR$_i$ $\downarrow$ \\
\midrule
\multirow{5}{*}{InstructBLIP}
  & \textit{base}         & 1.84 & 0.64 & 0.70 & 9.1 \\
  & VCD                 & 1.75 & 0.64 & 0.80 & 8.9 \\
  & M3ID                & 1.70 & 0.65 & 0.90 & \colorbox{SecondBest}{7.6} \\
  & OPERA (Beam)        & --   & --   & 16.6 & \colorbox{Best}{6.8} \\
  & \Ours               & \colorbox{Best}{2.03} & \colorbox{Best}{0.59} & \colorbox{Best}{0.70} & 8.3 \\
\midrule
\multirow{5}{*}{LLaVA-1.5}
  & \textit{base}         & 1.59 & 0.72 & 25.0 & 9.2 \\
  & VCD                 & 1.96 & 0.64 & 23.6 & 8.4 \\
  & M3ID                & \colorbox{SecondBest}{2.14} & \colorbox{SecondBest}{0.61} & 23.2 & \colorbox{Best}{7.3} \\
  & OPERA (Beam)        & 2.15 & \colorbox{Best}{0.54} & 45.1 & 22.3 \\
  & \Ours               & \colorbox{Best}{2.19} & 0.59 & \colorbox{Best}{22.1} & \colorbox{SecondBest}{7.8} \\
\bottomrule
\end{tabular}
}
\vspace{-2mm}
\caption{
\textbf{Results on free-form generative benchmarks.}
}
\label{tab:generative_benchmark}
\vspace{-2mm}
\end{table}

\begin{table*}[t]
    
    \centering
    \small
    \setlength\tabcolsep{4pt} 
    \scalebox{1}{
    \begin{tabular}{lllx{27}x{27}x{27}x{27}x{27}x{27}x{27}x{27}}
    \toprule
     & \multirow{2}{*}{\textbf{Setup}} & \multirow{2}{*}{\textbf{Method}} & \multicolumn{4}{c}{\textbf{InstructBLIP~\citep{dai2024instructblip}}} & \multicolumn{4}{c}{\textbf{LLaVA 1.5~\citep{liu2023visual}}}\\
    \arrayrulecolor{gray} \cmidrule(lr){4-7} \cmidrule(lr){8-11}
     &  &  & {{Acc.\up}} & {{Prec.\up}} & {{Rec.\up}} & {{F1\up}} & {{Acc.\up}} & {{Prec.\up}} & {{Rec.\up}} & {{F1\up}} \\
    \midrule
    \multirow{12}{*}{\rot{\textbf{\normalsize MS-COCO}}} & \multirow{4}{*}{Random} & \textit{base} &  81.53 & 82.71 & \colorbox{SecondBest}{79.73} & 81.19 & 83.77 & 92.31 & \colorbox{SecondBest}{73.67} & 81.94  \\
     &  & VCD & \colorbox{SecondBest}{82.03} & \colorbox{SecondBest}{83.77} & \colorbox{Best}{79.47} & \colorbox{SecondBest}{81.56} & \colorbox{Best}{85.43} & \colorbox{SecondBest}{93.25} & \colorbox{Best}{76.40} & \colorbox{Best}{83.99}  \\
     &  & \textbf{\Ours} & \colorbox{Best}{86.03} & \colorbox{Best}{95.53} & 75.60 & \colorbox{Best}{84.41} & \colorbox{SecondBest}{84.67} & \colorbox{Best}{97.88} & 70.87 & \colorbox{SecondBest}{82.21}
     \\
     \arrayrulecolor{gray!50}\cmidrule(lr){3-11}
      & \multirow{4}{*}{Popular} & \textit{base} &78.47&	77.73&	\colorbox{Best}{79.80} &	78.75& 82.57&	\colorbox{SecondBest}{89.62}&	\colorbox{SecondBest}{73.67}&	80.86 \\
     &  & VCD & \colorbox{SecondBest}{79.13}	& \colorbox{SecondBest}{78.94}	& \colorbox{SecondBest}{79.47} &	\colorbox{SecondBest}{79.20} &  \colorbox{SecondBest}{83.17}&	88.36&	\colorbox{Best}{76.40}&	\colorbox{Best}{81.94} \\
     &  & \textbf{\Ours} & \colorbox{Best}{84.27}	& \colorbox{Best}{91.45}&	75.60	&\colorbox{Best}{82.77}&  \colorbox{Best}{83.67}&	\colorbox{Best}{95.25}&	70.87	&\colorbox{SecondBest}{81.27}\\
     \arrayrulecolor{gray!50}\cmidrule(lr){3-11} 
      & \multirow{4}{*}{Adversarial} & \textit{base} & \colorbox{SecondBest}{77.43}&	76.09	&\colorbox{Best}{80.00}	&\colorbox{SecondBest}{78.00} & 79.77&	\colorbox{SecondBest}{83.85}	&\colorbox{SecondBest}{73.73}&	78.47 \\
     &  & VCD &  77.23	&\colorbox{SecondBest}{76.10}&	\colorbox{SecondBest}{79.40}&	77.72& \colorbox{SecondBest}{80.27}&	82.76&	\colorbox{Best}{76.47}	&\colorbox{SecondBest}{79.49}\\
     &  & \textbf{\Ours} & \colorbox{Best}{81.83}&	\colorbox{Best}{86.20}&	75.80	&\colorbox{Best}{80.67} &  \colorbox{Best}{81.83}&	\colorbox{Best}{90.99}&	70.67	&\colorbox{Best}{79.55}  \\
     \arrayrulecolor{gray}\midrule
     \multirow{12}{*}{\rot{\textbf{\normalsize A-OKVQA}}} & \multirow{4}{*}{Random} & \textit{base}& 81.33	&78.52	&\colorbox{SecondBest}{86.27}	&82.21 & 84.93&	\colorbox{SecondBest}{89.16}&	\colorbox{SecondBest}{79.53}&	84.07\\
     &  & VCD & \colorbox{SecondBest}{81.57}&	\colorbox{SecondBest}{78.78}&	\colorbox{Best}{86.40}&	\colorbox{SecondBest}{82.42}&  \colorbox{SecondBest}{85.53}&	87.64&	\colorbox{Best}{82.73}&	\colorbox{SecondBest}{85.12} \\
     &  & \textbf{\Ours} &  \colorbox{Best}{87.10}&	\colorbox{Best}{89.95}&	83.53&	\colorbox{Best}{86.62}& \colorbox{Best}{87.33}&	\colorbox{Best}{95.09}&	78.73&	\colorbox{Best}{86.14} \\
     \arrayrulecolor{gray!50}\cmidrule(lr){3-11} 
      & \multirow{4}{*}{Popular} & \textit{base} & 76.87&	72.69&	\colorbox{SecondBest}{86.07}&	78.82& 80.90&	\colorbox{SecondBest}{81.77}&	\colorbox{SecondBest}{79.53}&	80.64\\
     &  & VCD & \colorbox{SecondBest}{77.30}	&\colorbox{SecondBest}{73.10}&	\colorbox{Best}{86.40}&	\colorbox{SecondBest}{79.19}& \colorbox{SecondBest}{81.17}&	80.22&	\colorbox{Best}{82.73}&	\colorbox{SecondBest}{81.46} \\
     &  & \textbf{\Ours} & \colorbox{Best}{82.47}	&\colorbox{Best}{81.79}&	83.53&	\colorbox{Best}{82.65} &\colorbox{Best}{85.03}	&\colorbox{Best}{90.08}&	78.73&	\colorbox{Best}{84.03}   \\
     \arrayrulecolor{gray!50}\cmidrule(lr){3-11}
      & \multirow{4}{*}{Adversarial} & \textit{base} & 71.40&	66.67	&\colorbox{SecondBest}{85.60}	&74.96&  \colorbox{SecondBest}{74.80}&	\colorbox{SecondBest}{72.63}&	\colorbox{SecondBest}{79.60}&	75.95\\
     &  & VCD & \colorbox{SecondBest}{72.47}&	\colorbox{SecondBest}{67.39}	&\colorbox{Best}{87.07}	&\colorbox{SecondBest}{75.97}& \colorbox{SecondBest}{75.03}&	71.87&	\colorbox{Best}{82.27}&	\colorbox{SecondBest}{76.72} \\
     &  & \textbf{\Ours} & \colorbox{Best}{76.47}	&\colorbox{Best}{73.16}&	83.60&	\colorbox{Best}{78.03}& \colorbox{Best}{79.27}&	\colorbox{Best}{79.58}&	78.73&	\colorbox{Best}{79.16} \\
     \arrayrulecolor{gray}\midrule
     \multirow{12}{*}{\rot{\textbf{\normalsize GQA}}} & \multirow{4}{*}{Random} & \textit{base} & 80.57&	77.47&	\colorbox{SecondBest}{86.20}&	81.60& 84.80&	\colorbox{SecondBest}{87.88}&	\colorbox{SecondBest}{80.73}&	84.16 \\
     &  & VCD & \colorbox{SecondBest}{81.73}	&\colorbox{SecondBest}{79.02}	&\colorbox{Best}{86.40}&	\colorbox{SecondBest}{82.55}&  \colorbox{SecondBest}{85.63}&	86.89&	\colorbox{Best}{83.93}&	\colorbox{SecondBest}{85.38} \\
     &  & \textbf{\Ours} & \colorbox{Best}{85.30}&	\colorbox{Best}{88.57}&	81.07&	\colorbox{Best}{84.65}&   \colorbox{Best}{87.40}&	\colorbox{Best}{95.17}&	78.80&	\colorbox{Best}{86.21}\\
     \arrayrulecolor{gray!50}\cmidrule(lr){3-11}
      & \multirow{4}{*}{Popular} & \textit{base} &  74.67	&\colorbox{SecondBest}{70.17}&	\colorbox{SecondBest}{85.80}&	77.20  &\colorbox{SecondBest}{79.37}&	\colorbox{SecondBest}{78.59}&	\colorbox{SecondBest}{80.73}&	79.64\\
     &  & VCD & \colorbox{SecondBest}{74.63}	&69.94&	\colorbox{Best}{86.40}&	\colorbox{SecondBest}{77.30}& 78.73&	76.03&	\colorbox{Best}{83.93}&	\colorbox{SecondBest}{79.78} \\
     &  & \textbf{\Ours} & \colorbox{Best}{80.63}	&\colorbox{Best}{80.37}&	81.07&	\colorbox{Best}{80.72}&  \colorbox{Best}{83.33}&	\colorbox{Best}{86.66}	&78.80&	\colorbox{Best}{82.54}\\
     \arrayrulecolor{gray!50}\cmidrule(lr){3-11}
      & \multirow{4}{*}{Adversarial} & \textit{base} &\colorbox{SecondBest}{72.63}	&\colorbox{SecondBest}{67.78}&	\colorbox{Best}{86.27}&	\colorbox{SecondBest}{75.92} &76.00&	\colorbox{SecondBest}{74.13}&	\colorbox{SecondBest}{79.87}&	76.89\\
     &  & VCD &  71.93&	67.21&	\colorbox{SecondBest}{85.67}&	75.32& \colorbox{SecondBest}{76.40}&	72.76&	\colorbox{Best}{84.40}&	\colorbox{SecondBest}{78.15} \\
     &  & \textbf{\Ours} & \colorbox{Best}{77.60}&	\colorbox{Best}{75.91}&	80.87&	\colorbox{Best}{78.31}& \colorbox{Best}{80.37}&	\colorbox{Best}{81.52}&	78.53&	\colorbox{Best}{80.00}\\
    \bottomrule
    \end{tabular}
    }
    \vspace{-2mm}
    \caption{
        \textbf{POPE~\citep{li2023evaluating} results with one-word constraint.}
        We use the instruction "Please answer in one word." at the end of the query text.
    }
    \vspace{-2mm}
    \label{tab:POPE_one_word}
\end{table*}

\subsection{POPE~\citep{li2023evaluating} with Single-Word Constraint}
As shown in \cref{tab:POPE_one_word}, we see that imposing a one-word response constraint on LVLMs leads to notable changes in performance compared to~\cref{tab:POPE}.
Despite the change in query setup, \Ours shows the best performance on the POPE benchmark. 
Specifically, precision and recall vary significantly in the COCO random setup comparing scenarios with and without the instruction, "Please answer this question with one word."
To mitigate these impacts and better evaluate discriminative capabilities, we designed experiments that allow the LVLMs to freely make judgments and provide explanations for these judgments rather than restricting them to answers in one word.

\begin{table*}[t!]
\centering
\small 
\setlength\tabcolsep{2pt} 
\scalebox{0.9}{
\def\arraystretch{1.2}
\newcolumntype{K}{!{\color{white}\ }c}
\begin{tabular}{ccKKKKKKKK}

\toprule
\multirow{2}{*}{Task}        & \multirow{2}{*}{Category}                                          & \multicolumn{4}{c}{LLaVA 1.5~\citep{liu2023visual}}     & \multicolumn{4}{c}{InstructBLIP~\citep{dai2024instructblip}}  \\ \arrayrulecolor{gray} \cmidrule(lr){3-6} \cmidrule(lr){7-10} 
                             &                                                                    & \textit{base}   & VCD    & M3ID   & \Ours & \textit{base}   & VCD    & M3ID   & \Ours \\ \midrule  \vspace{0.1cm}

\multirow{14}{*}{\rot{\textbf{Perception}}}
& Existence    &	\shortstack{173.57\\ $_{(\pm8.16)}$}&	\shortstack{172.14\\ $_{(\pm8.09)}$}&	\colorbox{SecondBest}{\shortstack{178.33\\ $_{(\pm6.83)}$}}&	\colorbox{Best}{\shortstack{189.29\\ $_{(\pm1.89)}$}}  & \shortstack{170.19\\ $_{(\pm11.12)}$}&	\shortstack{172.62\\ $_{(\pm8.92)}$}&	\colorbox{SecondBest}{\shortstack{173.89\\ $_{(\pm10.52)}$}}&	\colorbox{Best}{\shortstack{184.76\\ $_{(\pm5.56)}$}}\\ \vspace{0.1cm} 
                             & Count                                                              & \colorbox{SecondBest}{\shortstack{110.00\\ $_{(\pm15.82)}$}}&	\colorbox{Best}{\shortstack{117.14\\ $_{(\pm8.76)}$}}&	\shortstack{107.22\\ $_{(\pm14.78)}$}&	\shortstack{104.76\\ $_{(\pm11.66)}$}&	\shortstack{89.52\\ $_{(\pm11.04)}$}&	\colorbox{Best}{\shortstack{98.33\\ $_{(\pm15.99)}$}}&	\colorbox{SecondBest}{\shortstack{89.72\\ $_{(\pm13.44)}$}}&	\shortstack{82.85\\ $_{(\pm12.16)}$}
  \\ \vspace{0.1cm}
                             & Position                                                           & \shortstack{100.47\\ $_{(\pm18.78)}$}&	\colorbox{SecondBest}{\shortstack{103.33\\ $_{(\pm20.56)}$}}&	\shortstack{96.39\\ $_{(\pm5.52)}$}&	\colorbox{Best}{\shortstack{106.19\\ $_{(\pm13.93)}$}}&	\shortstack{67.62\\ $_{(\pm14.04)}$}&	\shortstack{71.90\\ $_{(\pm13.42)}$}&	\colorbox{SecondBest}{\shortstack{72.72\\ $_{(\pm14.77)}$}}&	\colorbox{Best}{\shortstack{74.76\\ $_{(\pm6.19)}$}}
  \\ \vspace{0.1cm}
                             & Color                                                              & \shortstack{125.24\\ $_{(\pm15.91)}$}&	\shortstack{119.52\\ $_{(\pm8.58)}$}&	\colorbox{SecondBest}{\shortstack{127.50\\ $_{(\pm8.28)}$}}&	\colorbox{Best}{\shortstack{127.86\\ $_{(\pm9.13)}$}}&	\shortstack{114.76\\ $_{(\pm9.60)}$}&	\colorbox{SecondBest}{\shortstack{117.14\\ $_{(\pm10.70)}$}}&	\shortstack{110.56\\ $_{(\pm7.20)}$}&	\colorbox{Best}{\shortstack{131.43\\ $_{(\pm4.76)}$}}
 \\ \vspace{0.1cm}
                             & Posters                                                            & \shortstack{132.31\\ $_{(\pm6.73)}$}&	\colorbox{SecondBest}{\shortstack{135.54\\ $_{(\pm3.61)}$}}&	\shortstack{132.82\\ $_{(\pm7.94)}$}&	\colorbox{Best}{\shortstack{150.85\\ $_{(\pm6.49)}$}}&	\shortstack{114.97\\ $_{(\pm6.25)}$}&	\colorbox{SecondBest}{\shortstack{129.08\\ $_{(\pm6.97)}$}}&	\shortstack{114.46\\ $_{(\pm6.97)}$}&	\colorbox{Best}{\shortstack{145.92\\ $_{(\pm2.41)}$}}
 \\ \vspace{0.1cm}
                             & Celebrity                                                          &   \shortstack{114.56\\ $_{(\pm6.45)}$}&	\colorbox{SecondBest}{\shortstack{118.09\\ $_{(\pm7.69)}$}}&	\shortstack{113.38\\ $_{(\pm0.21)}$}&	\colorbox{Best}{\shortstack{125.59\\ $_{(\pm2.50)}$}}&	\shortstack{113.38\\ $_{(\pm3.95)}$}&	\colorbox{Best}{\shortstack{123.82\\ $_{(\pm4.99)}$}}&	\shortstack{114.12\\ $_{(\pm2.91)}$}&	\colorbox{SecondBest}{\shortstack{120.29\\ $_{(\pm7.90)}$}}
 \\ \vspace{0.1cm}
                             & Scene                                                              & \shortstack{149.13\\ $_{(\pm0.53)}$}&	\shortstack{150.00\\ $_{(\pm3.54)}$}&	\colorbox{SecondBest}{\shortstack{156.63\\ $_{(\pm1.59)}$}}&	\colorbox{Best}{\shortstack{162.00\\ $_{(\pm1.06)}$}}&	\shortstack{140.50\\ $_{(\pm0.71)}$}&	\shortstack{136.50\\ $_{(\pm10.25)}$}&	\colorbox{SecondBest}{\shortstack{141.00\\ $_{(\pm1.06)}$}}&	\colorbox{Best}{\shortstack{150.38\\ $_{(\pm3.36)}$}}
 \\ \vspace{0.1cm}
                             & Landmark                                                           & \shortstack{138.25\\ $_{(\pm4.95)}$}&	\colorbox{SecondBest}{\shortstack{140.75\\ $_{(\pm4.95)}$}}&	\shortstack{135.13\\ $_{(\pm4.77)}$}&	\colorbox{Best}{\shortstack{142.38\\ $_{(\pm0.53)}$}}&	\shortstack{98.50\\ $_{(\pm0.35)}$}&	\colorbox{Best}{\shortstack{110.75\\ $_{(\pm4.24)}$}}&	\colorbox{SecondBest}{\shortstack{103.25\\ $_{(\pm6.72)}$}}&	\shortstack{99.25\\ $_{(\pm0.35)}$}
 \\ \vspace{0.1cm}
                             & Artwork                                                            & \colorbox{SecondBest}{\shortstack{97.50\\ $_{(\pm2.83)}$}}&	\shortstack{95.25\\ $_{(\pm4.24)}$}&	\shortstack{89.38\\ $_{(\pm3.36)}$}&	\colorbox{Best}{\shortstack{101.00\\ $_{(\pm7.42)}$}}&	\shortstack{110.38\\ $_{(\pm4.42)}$}&	\colorbox{SecondBest}{\shortstack{113.00\\ $_{(\pm3.54)}$}}&	\shortstack{110.13\\ $_{(\pm6.89)}$}&	\colorbox{Best}{\shortstack{123.38\\ $_{(\pm2.30)}$}}
 \\ 
                             & OCR                                                                & \shortstack{91.25\\ $_{(\pm19.45)}$}&	\colorbox{SecondBest}{\shortstack{101.25\\ $_{(\pm1.77)}$}}&	\shortstack{96.25\\ $_{(\pm15.91)}$}&	\colorbox{Best}{\shortstack{143.75\\ $_{(\pm5.3)}$}}&	\colorbox{SecondBest}{\shortstack{87.50\\ $_{(\pm21.21)}$}}&	\colorbox{Best}{\shortstack{91.25\\ $_{(\pm8.84)}$}}&	\shortstack{85.00\\ $_{(\pm10.61)}$}&	\shortstack{68.75\\ $_{(\pm5.3)}$}
  \\ \arrayrulecolor{gray} \cmidrule(lr){1-10} \vspace{0.1cm}
\multirow{6}{*}{\rot{\textbf{Recognition}}}
 & \shortstack{Commonsense\\Reasoning} 
 
 &\colorbox{SecondBest}{\shortstack{100.36\\ $_{(\pm2.53)}$}}&	\shortstack{96.79\\ $_{(\pm5.56)}$}&	\shortstack{87.14\\ $_{(\pm12.12)}$}&	\colorbox{Best}{\shortstack{102.86\\ $_{(\pm7.07)}$}}&	\shortstack{96.43\\ $_{(\pm1.01)}$}&	\colorbox{Best}{\shortstack{107.14\\ $_{(\pm8.08)}$}}&	\shortstack{99.64\\ $_{(\pm2.53)}$}&	\colorbox{SecondBest}{\shortstack{101.79\\ $_{(\pm6.57)}$}}
 \\ \vspace{0.1cm}
                             & \shortstack{Numerical\\Calculation}    
                             
                             & \colorbox{Best}{\shortstack{80.00\\ $_{(\pm7.07)}$}}&	\shortstack{66.25\\ $_{(\pm8.84)}$}&	\colorbox{SecondBest}{\shortstack{76.25\\ $_{(\pm12.37)}$}}&	\shortstack{65.00\\ $_{(\pm14.14)}$}&	\shortstack{68.75\\ $_{(\pm1.77)}$}&	\shortstack{66.25\\ $_{(\pm15.91)}$}&	\colorbox{SecondBest}{\shortstack{71.25\\ $_{(\pm22.98)}$}}&	\colorbox{Best}{\shortstack{73.75\\ $_{(\pm5.30)}$}}
  \\ \vspace{0.1cm}
                             & \shortstack{Text\\Translation}         
                             
                             & \shortstack{75.00\\ $_{(\pm3.54)}$}&	\colorbox{Best}{\shortstack{86.25\\ $_{(\pm22.98)}$}}&	\shortstack{65.00\\ $_{(\pm14.14)}$}&	\colorbox{SecondBest}{\shortstack{77.50\\ $_{(\pm17.68)}$}}&	\shortstack{63.75\\ $_{(\pm5.3)}$}&	\colorbox{Best}{\shortstack{91.25\\ $_{(\pm1.77)}$}}&	\shortstack{53.75\\ $_{(\pm5.3)}$}&	\colorbox{SecondBest}{\shortstack{86.25\\ $_{(\pm1.77)}$}}
  \\ 
                             & \shortstack{Code\\Reasoning}           &
                             
                            \shortstack{62.50\\ $_{(\pm10.61)}$}&	\shortstack{61.25\\ $_{(\pm1.77)}$}&	\colorbox{Best}{\shortstack{71.25\\ $_{(\pm15.91)}$}}&	\colorbox{Best}{\shortstack{71.25\\ $_{(\pm5.30)}$}}&	\shortstack{73.75\\ $_{(\pm5.30)}$}&	\shortstack{57.50\\ $_{(\pm0.00)}$}&	\colorbox{Best}{\shortstack{81.25\\ $_{(\pm1.77)}$}}&	\colorbox{SecondBest}{\shortstack{76.25\\ $_{(\pm5.3)}$}}
 \\
                     \bottomrule
\end{tabular}
}
\vspace{-2mm}
\caption{\textbf{Results on MME-Fullset~\citep{fu2024mme}.}}
\vspace{-2mm}
\label{tab:MME_full_apendix}
\end{table*}

\begin{table*}[t!]

\centering
\small 
\setlength\tabcolsep{2pt} 
\scalebox{1}{
\def\arraystretch{1.2}
\newcolumntype{K}{!{\color{white}\ }c}
\begin{tabular}{cKKKKKKKK}

\toprule
 \multirow{2}{*}{Category}                                          & \multicolumn{4}{c}{LLaVA 1.5~\citep{liu2023visual}}     & \multicolumn{4}{c}{InstructBLIP~\citep{dai2024instructblip}}  \\ \arrayrulecolor{gray} \cmidrule(lr){2-5} \cmidrule(lr){6-9}                                                                     & \textit{base}   & VCD    & M3ID   & \Ours & \textit{base}   & VCD    & M3ID   & \Ours \\ \midrule  \vspace{0.1cm}

 Existence    &	\colorbox{SecondBest}{\shortstack{68.55\\ $_{(\pm0.21)}$}}&	\shortstack{67.15\\ $_{(\pm1.91)}$}&	{\shortstack{68.50\\ $_{(\pm0.14)}$}}&	\colorbox{Best}{\shortstack{75.35\\ $_{(\pm0.21)}$}}  & \shortstack{72.05\\ $_{(\pm0.49)}$}&	\colorbox{SecondBest}{\shortstack{73.20\\ $_{(\pm1.27)}$}}&	{\shortstack{72.95\\ $_{(\pm0.21)}$}}&	\colorbox{Best}{\shortstack{81.35\\ $_{(\pm0.07)}$}}\\ \vspace{0.1cm} 
                             Attribute                                                              & {\shortstack{67.85\\ $_{(\pm0.49)}$}}&	\colorbox{SecondBest}{\shortstack{69.50\\ $_{(\pm1.27)}$}}&	\shortstack{68.20\\ $_{(\pm0.42)}$}&	\colorbox{Best}{\shortstack{69.80\\ $_{(\pm0.85)}$}}&	\shortstack{68.40\\ $_{(\pm0.14)}$}&	\colorbox{SecondBest}{\shortstack{69.90\\ $_{(\pm0.14)}$}}&	{\shortstack{69.15\\ $_{(\pm0.92)}$}}&	\colorbox{Best}{\shortstack{70.80\\ $_{(\pm1.56)}$}}
  \\ \vspace{0.1cm}
                             State                                                           & \shortstack{65.55\\ $_{(\pm0.35)}$}&	\colorbox{SecondBest}{\shortstack{67.80\\ $_{(\pm0.28)}$}}&	\shortstack{65.75\\ $_{(\pm0.64)}$}&	\colorbox{Best}{\shortstack{68.40\\ $_{(\pm1.70)}$}}&	\shortstack{70.55\\ $_{(\pm0.64)}$}&	\colorbox{SecondBest}{\shortstack{72.40\\ $_{(\pm0.00)}$}}&	{\shortstack{70.70\\ $_{(\pm0.85)}$}}&	\colorbox{Best}{\shortstack{72.85\\ $_{(\pm1.77)}$}}
  \\ \vspace{0.1cm}
                             Number                                                             & \colorbox{Best}{\shortstack{69.05\\ $_{(\pm0.78)}$}}&	\shortstack{68.50\\ $_{(\pm2.40)}$}&	\colorbox{SecondBest}{\shortstack{68.95\\ $_{(\pm0.92)}$}}&	{\shortstack{67.10\\ $_{(\pm1.84)}$}}&	\colorbox{SecondBest}{\shortstack{60.90\\ $_{(\pm0.00)}$}}&	{\shortstack{60.70\\ $_{(\pm0.85)}$}}&	\colorbox{Best}{\shortstack{61.80\\ $_{(\pm0.71)}$}}&	{\shortstack{60.85\\ $_{(\pm0.49)}$}}
 \\ \vspace{0.1cm}
                             Action                                                            & \shortstack{78.50\\ $_{(\pm3.96)}$}&	\colorbox{SecondBest}{\shortstack{81.90\\ $_{(\pm3.39)}$}}&	\shortstack{81.50\\ $_{(\pm1.84)}$}&	\colorbox{Best}{\shortstack{84.50\\ $_{(\pm3.25)}$}}&	\shortstack{74.95\\ $_{(\pm2.05)}$}&	\colorbox{SecondBest}{\shortstack{79.05\\ $_{(\pm2.62)}$}}&	\shortstack{78.70\\ $_{(\pm1.27)}$}&	\colorbox{Best}{\shortstack{85.20\\ $_{(\pm2.40)}$}}
 \\ \vspace{0.1cm}
                             Relation                                                          &   \shortstack{58.80\\ $_{(\pm4.10)}$}&	{\shortstack{57.75\\ $_{(\pm0.07)}$}}&	\colorbox{SecondBest}{\shortstack{59.70\\ $_{(\pm3.39)}$}}&	\colorbox{Best}{\shortstack{60.50\\ $_{(\pm0.14)}$}}&	\shortstack{56.05\\ $_{(\pm1.63)}$}&	\colorbox{Best}{\shortstack{58.00\\ $_{(\pm1.41)}$}}&	\colorbox{SecondBest}{\shortstack{57.00\\ $_{(\pm1.98)}$}}&	{\shortstack{54.65\\ $_{(\pm2.76)}$}}
 \\
                     \bottomrule
\end{tabular}
}
\vspace{-2mm}
\caption{\textbf{Results on AMBER discriminative tasks~\citep{wang2023llm}.}}
\vspace{-2mm}
\label{tab:AMBER_discriminative}
\end{table*}

\subsection{Detailed Results on MME-Fullset}
The detailed results on MME-Fullset are provided in~\cref{tab:MME_full_apendix}. \Ours demonstrates substantial improvements in both LLaVA-1.5 and InstructBLIP across a wide range of perception and recognition tasks. These findings highlight the capability of \Ours to effectively handle diverse tasks, extending beyond hallucination mitigation, and suggest its potential to enhance the ability of LVLMs to accurately interpret and analyze visual information and query text appropriately.

\subsection{Detailed Results on AMBER Discriminative Tasks}
\label{sec:appendix_mme_fullset}
\cref{tab:AMBER_discriminative} presents the performance of the discriminative task on the AMBER benchmark across different categories. The discriminative task in the AMBER benchmark is divided into six categories: 'Existence', 'Attribute', 'State', 'Number', 'Action', and 'Relation', to evaluate the model's performance. For most categories, except for a few, both the LLaVA-1.5 and InstructBLIP models show performance improvements due to the applied \Ours.

\begin{figure*}[t!]
    \centering
    \vspace{-1mm}
    \begin{minipage}{\textwidth}
        \includegraphics[width=.94\textwidth]{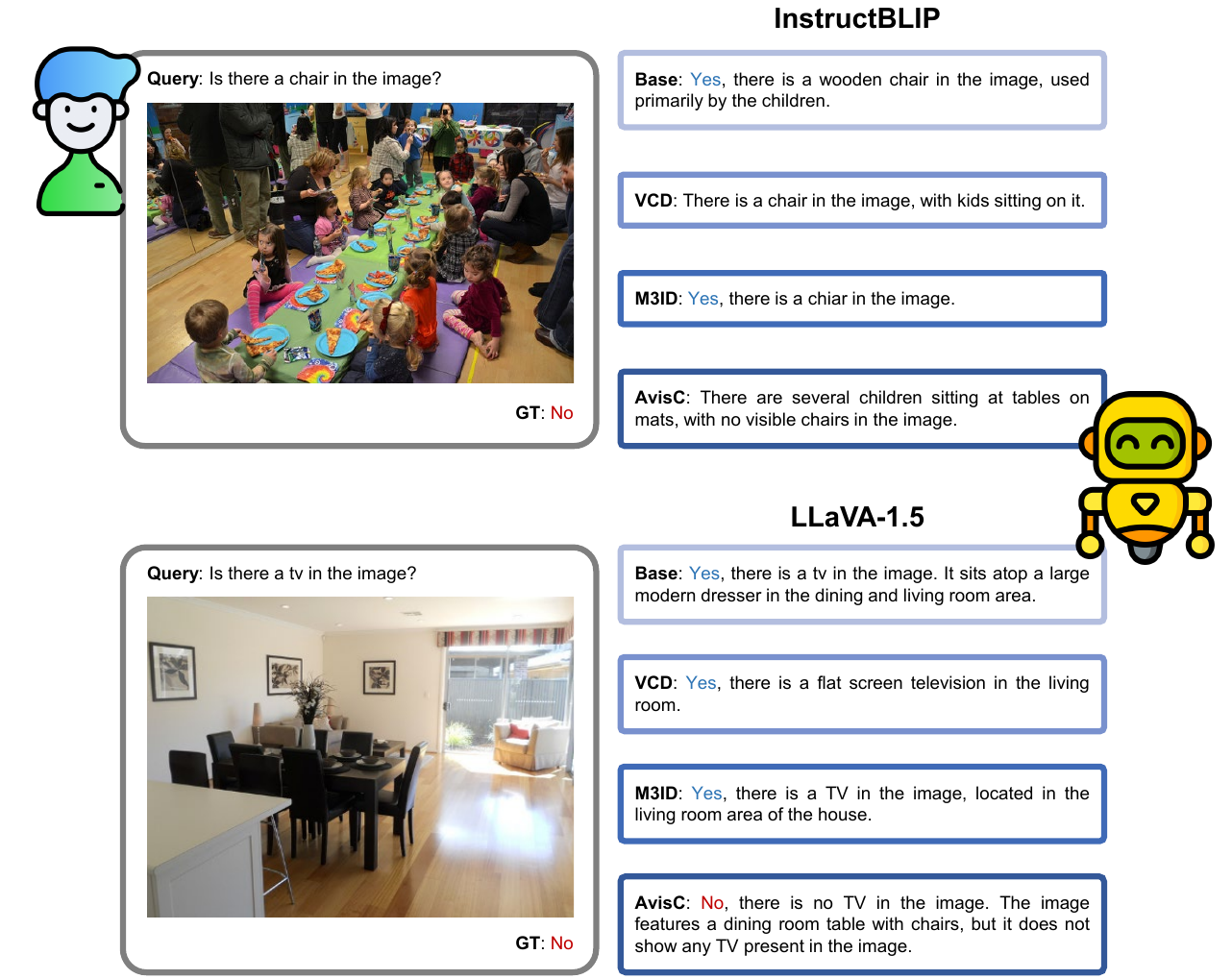}
    \end{minipage}
    \begin{minipage}{\textwidth}
        \includegraphics[width=.94\textwidth]{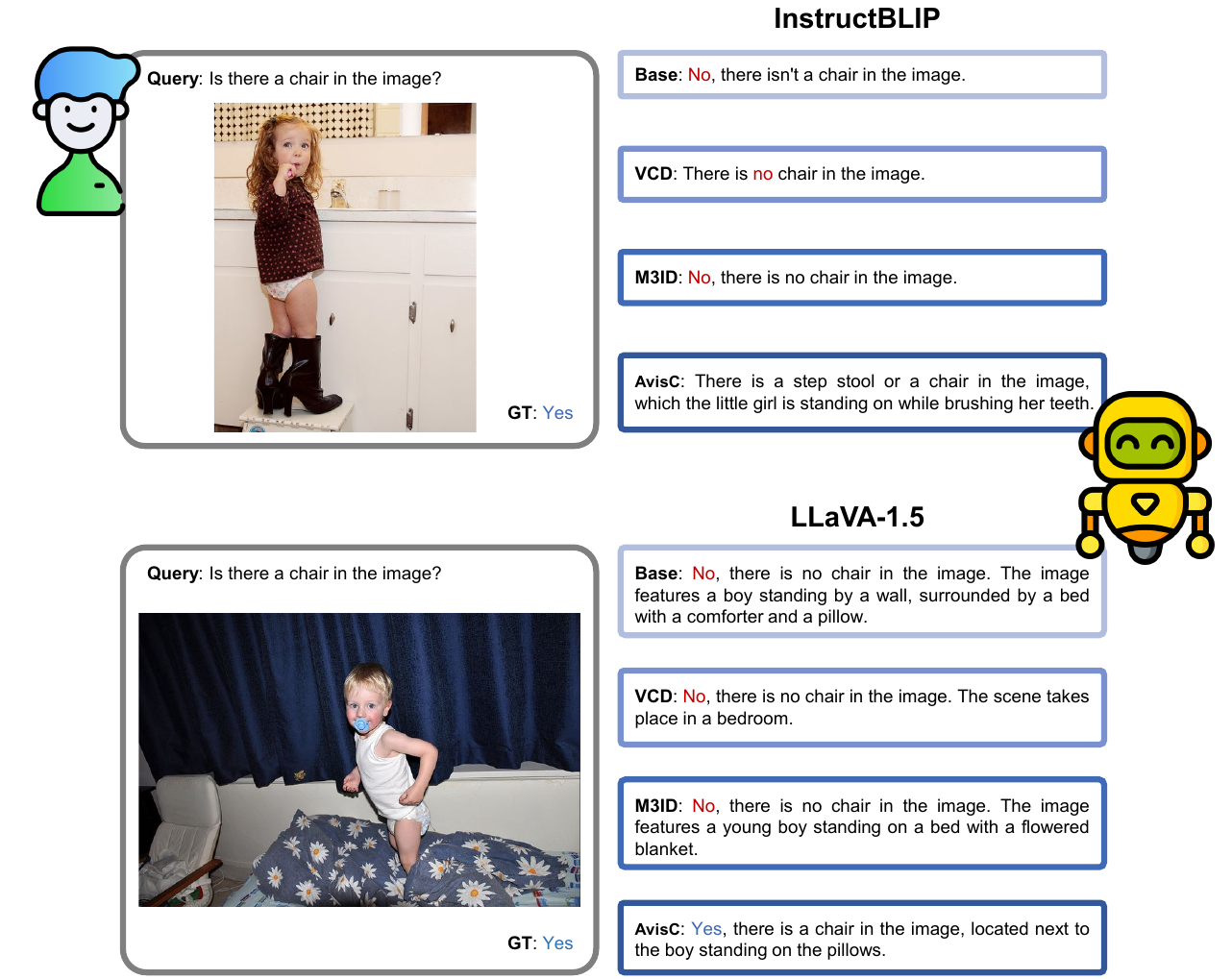}
    \end{minipage}
    \caption{
        \textbf{Qualitative examples on POPE~\citep{li2023evaluating}.}
    }%
    \label{fig:qualitative_pope}
\end{figure*}
\begin{figure*}[t!]
    \centering
    \vspace{-1mm}
    \begin{minipage}{.94\textwidth}
        \includegraphics[width=\textwidth]{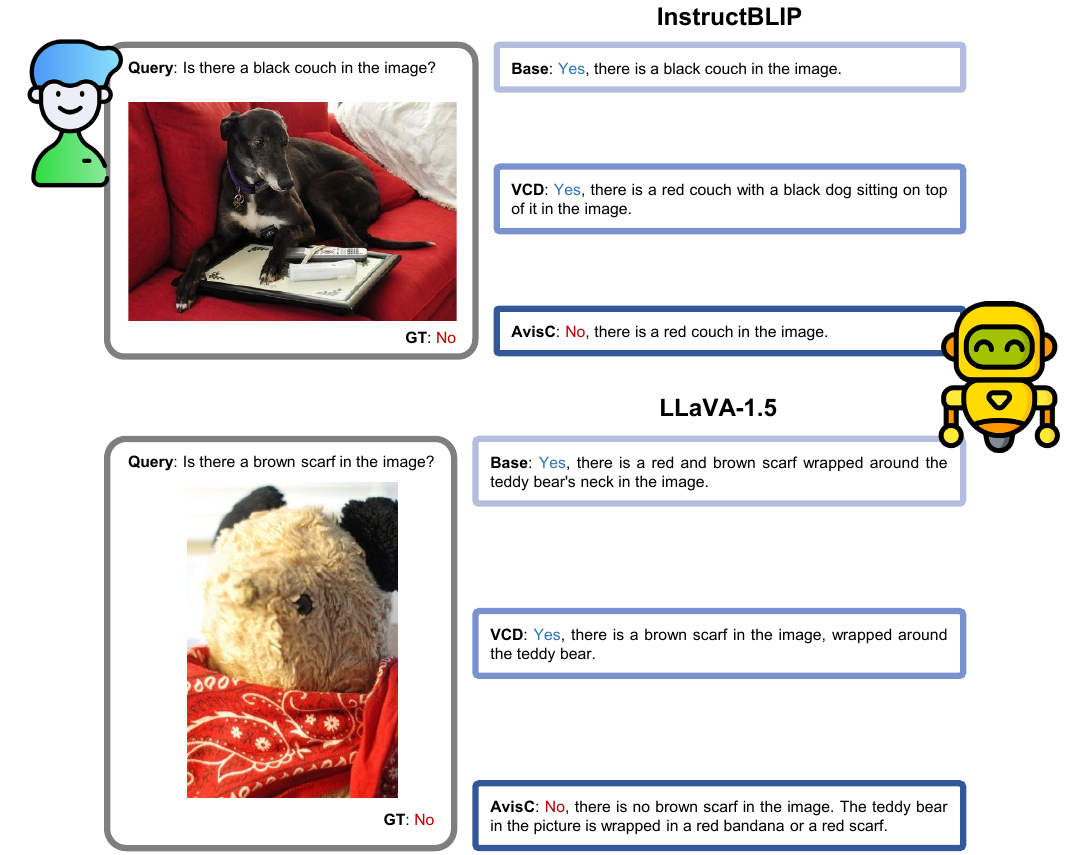}
    \end{minipage}
    \begin{minipage}{.94\textwidth}
        \includegraphics[width=\textwidth]{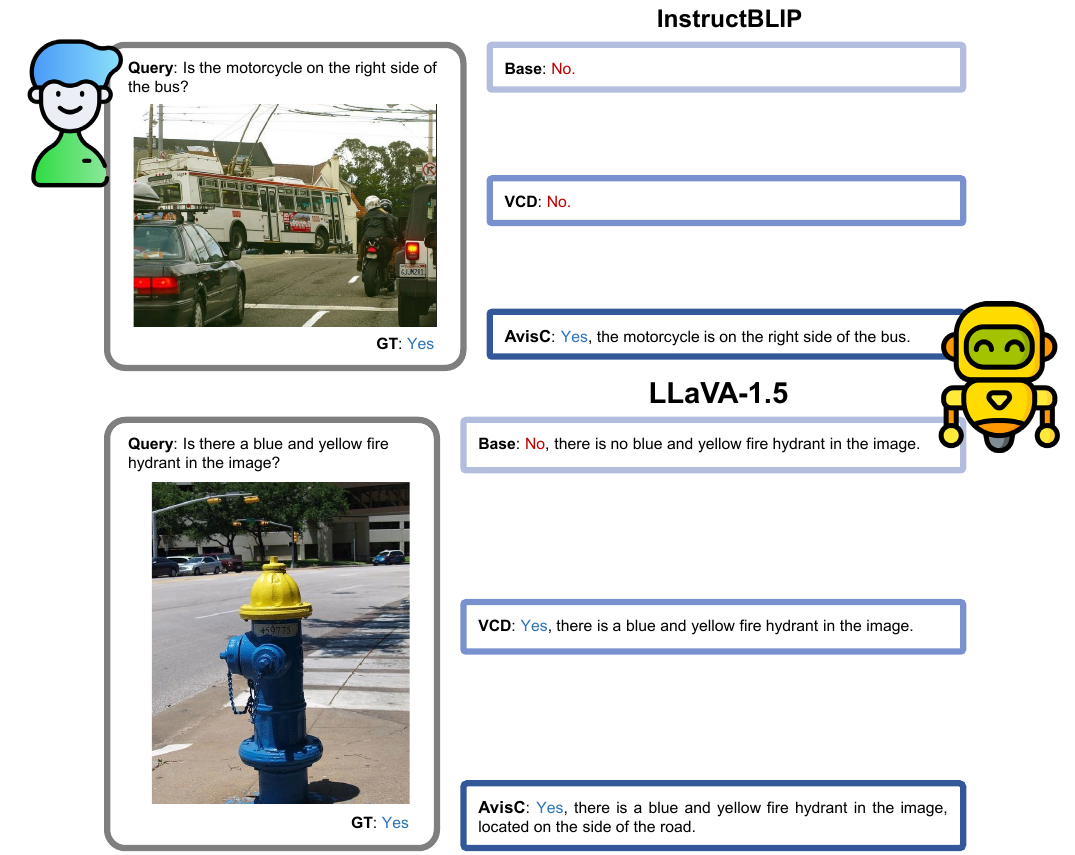}
    \end{minipage}
    \caption{
        \textbf{Qualitative examples on MME~\citep{fu2024mme}.}
    }%
    \label{fig:qualitative_mme}
\end{figure*}
\begin{figure*}[t!]
    \centering
    \vspace{-1mm}
    \begin{minipage}{\textwidth}
        \includegraphics[width=\textwidth]{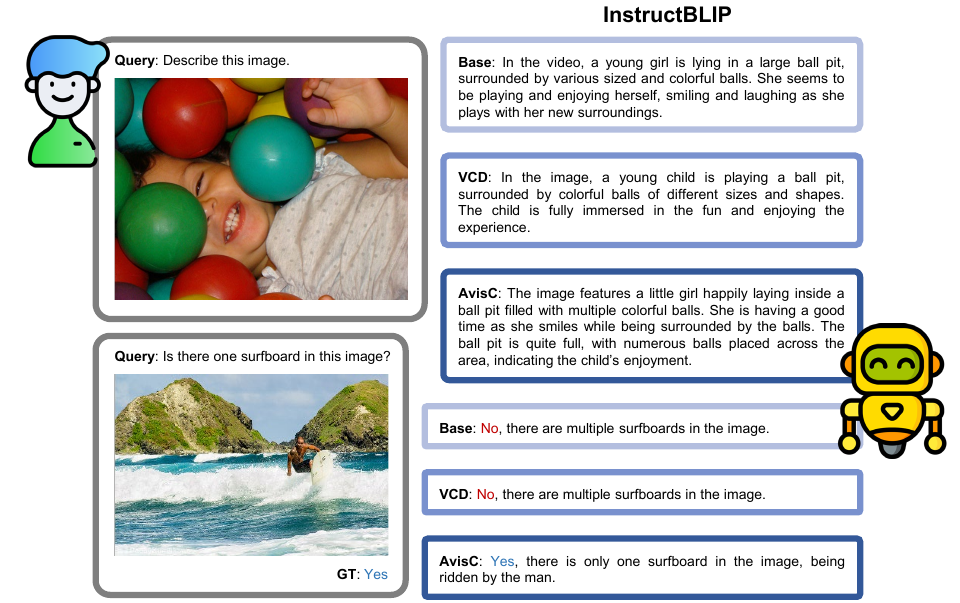}
    \end{minipage}
    \begin{minipage}{\textwidth}
        \includegraphics[width=\textwidth]{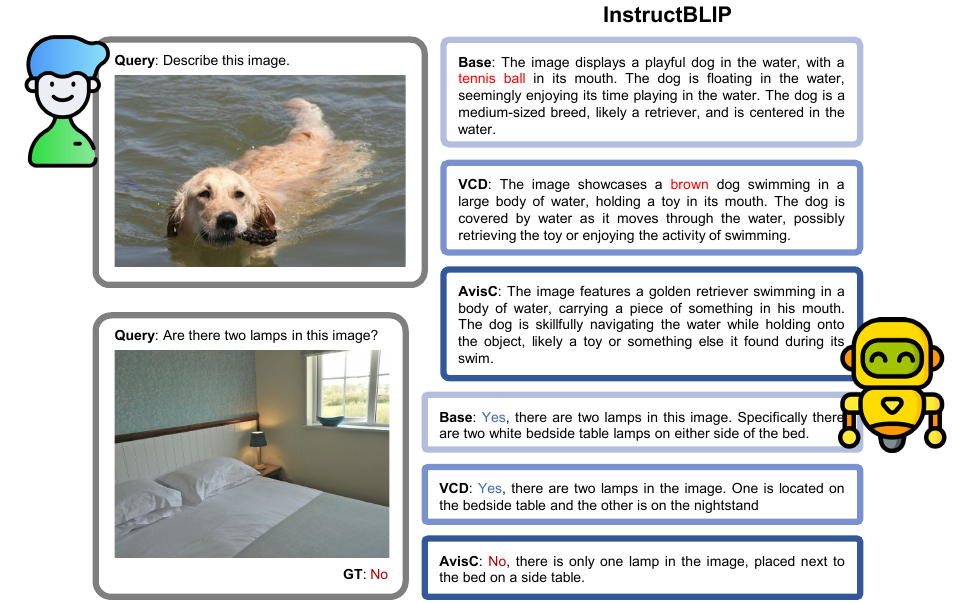}
    \end{minipage}
    \caption{
        \textbf{Qualitative examples of InstructBLIP~\citep{dai2024instructblip} on AMBER~\citep{wang2023llm}.}
    }%
    \label{fig:qualitative_amber_instructblip}
\end{figure*}
\begin{figure*}[t!]
    \centering
    \vspace{-1mm}
    \begin{minipage}{\textwidth}
    \includegraphics[width=\textwidth]{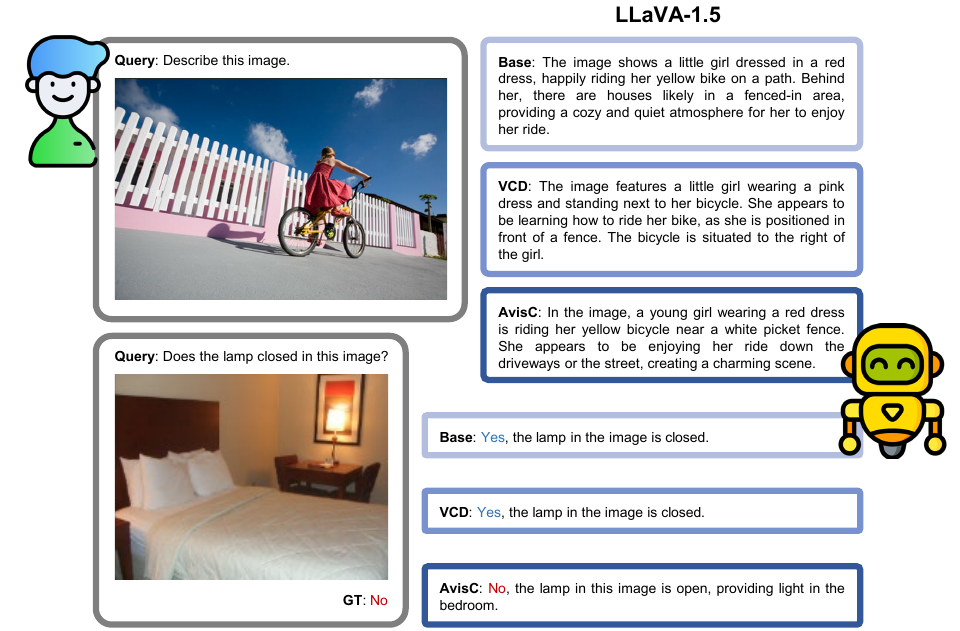}
    \end{minipage}
    \begin{minipage}{\textwidth}
        \includegraphics[width=\textwidth]{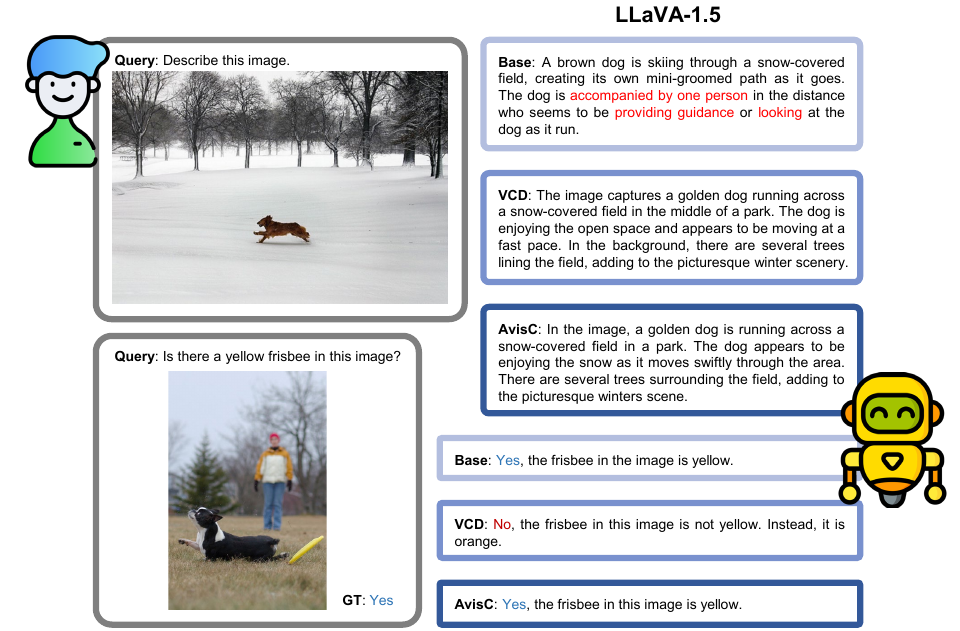}
    \end{minipage}
    \caption{
        \textbf{Qualitative examples of LLaVA-1.5~\citep{liu2023improved} on AMBER~\citep{wang2023llm}.}
    }%
    \label{fig:qualitative_amber_llava}
\end{figure*}
\begin{figure*}[t!]
    \centering
    \vspace{-1mm}
    \includegraphics[width=\textwidth]{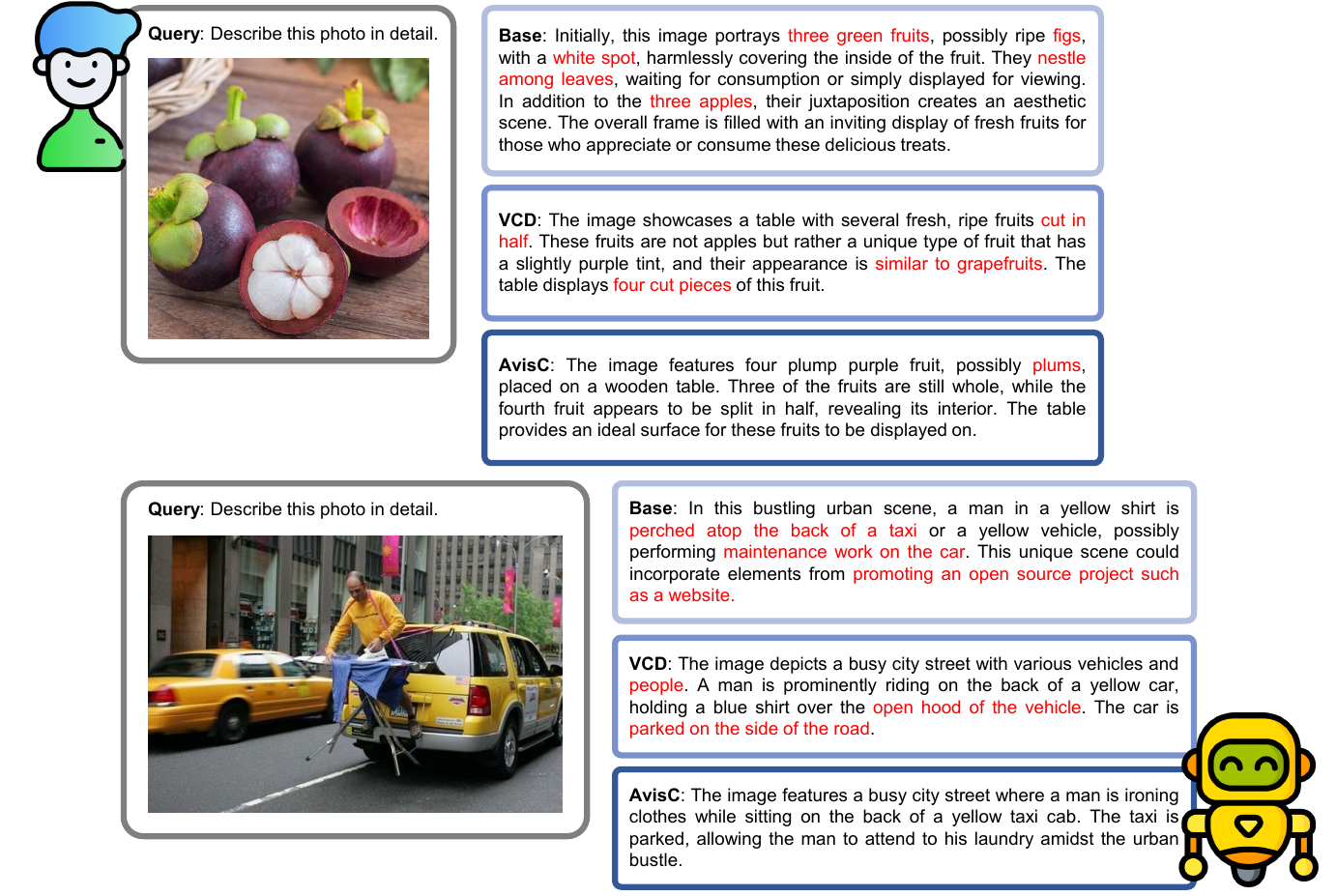}
    \caption{
        \textbf{Response comparison on LLaVA-Bench~\citep{liu2023visual}.}
        Hallucinations are colored in {\color{red}{red}}.
        \Ours demonstrates a robust understanding of images and reduces hallucinations in responses.
    }%
    \label{fig:qualitative_llava_bench}
\end{figure*}

\section{Comparison with "Vision Transformers Need Registers"~\citep{darcet2023vision}}
\label{sec:appendix_discussion}

\paragraph{Summary of~\citep{darcet2023vision}.}
\cite{darcet2023vision} identify artifacts in vision transformer feature maps—specifically, “high-norm outlier tokens” that concentrate attention in redundant background areas. These tokens capture significant global information despite lacking local details, leading to poor performance in tasks requiring spatial precision. Notably, when additional memory (register tokens) is introduced, these artifacts vanish.

\paragraph{Differences from blind tokens.}
While both high-norm outlier tokens and our \emph{blind tokens} exhibit unusually high attention weights in seemingly irrelevant regions, key differences exist:
\begin{itemize}
\item \textbf{Source of Attention:} High-norm tokens are computed within vision transformer layers, whereas our blind tokens are derived from the LLM’s attention in LVLMs (e.g., Vicuna-7B in LLaVA-1.5-7B), with differences in masking strategies.
\item \textbf{Task and Architecture:} Vision transformers are optimized for dense prediction tasks, and the emergence of high-norm tokens is sensitive to the training regime (e.g., DINOv1 vs. DINOv2). In contrast, LVLMs integrate a visual encoder with an LLM via an auto-regressive prediction scheme for image-based Q\&A tasks.
\item \textbf{Domain:} LVLMs project image tokens into an LLM space, altering attention dynamics compared to pure vision transformers.
\end{itemize}
Our experiments show a moderate correlation between high-norm and blind tokens (with $P(\text{blind token}\mid\text{high-norm token}) = 40.38\%$ and $P(\text{high-norm token}\mid\text{blind token}) = 31.27\%$), suggesting shared underlying properties despite their differences. Additionally, blind tokens tend to appear at the beginning and end of the image token sequence—a pattern not clearly observed for high-norm tokens in vision transformers.

\paragraph{On reducing dependency on blind tokens.}
Although high-norm tokens in \citep{darcet2023vision} encode global information, our findings indicate that blind tokens in LVLMs often lack query-relevant details. As shown in \cref{fig:motivation}, the essential information is typically captured by non-blind tokens. Therefore, reducing the influence of blind tokens via our contrastive decoding scheme---while enhancing the role of non-blind tokens---effectively mitigates hallucinations without sacrificing critical information.

\section{Qualitative Results}
We provide qualitative results on all benchmarks (POPE~\citep{li2023evaluating}, MME~\citep{fu2024mme}, AMBER~\citep{wang2023llm}, and LLaVA-Bench~\citep{liu2023visual}) in~\cref{fig:qualitative_pope,fig:qualitative_mme,fig:qualitative_amber_instructblip,fig:qualitative_amber_llava,}.
These highlight the differences between sentences generated by standard decoding (Base), VCD~\citep{leng2023mitigating}, and those produced by \Ours.
The results demonstrate the effectiveness of \Ours in dealing with a variety of challenging visual contexts.
Base and VCD often generate descriptions that include errors or hallucinations where elements not present in the image are described.
In contrast, \Ours helps counteract these hallucinations, generating sentences that reflect a more accurate comprehension of the image.

\end{document}